\documentclass[10pt,twocolumn,letterpaper]{article}
\usepackage{iccv}

\usepackage[toc,page]{appendix}

\newcommand{\resumetocwriting}{%
  \addtocontents{toc}{\protect\setcounter{tocdepth}{\arabic{tocdepth}}}}

\usepackage{times}
\usepackage{epsfig}
\usepackage{graphicx}
\usepackage{amsmath}
\usepackage{amssymb}
\usepackage{booktabs}
\usepackage{algpseudocode}
\usepackage{multirow}
\usepackage{adjustbox}
\usepackage{subfig}
\usepackage[dvipsnames]{xcolor}
\usepackage[ruled,vlined]{algorithm2e}
\usepackage[none]{hyphenat}
\usepackage{enumitem}
\usepackage{relsize}

\definecolor{commentcolor}{RGB}{110,154,155}   
\newcommand{\PyComment}[1]{\ttfamily\textcolor{commentcolor}{\# #1}}  
\newcommand{\PyCode}[1]{\ttfamily\textcolor{black}{#1}} 

\counterwithin*{equation}{section}

\usepackage{xr}
\makeatletter

\newcommand*{\addFileDependency}[1]{
\typeout{(#1)}
\@addtofilelist{#1}
\IfFileExists{#1}{}{\typeout{No file #1.}}
}\makeatother

\newcommand*{\myexternaldocument}[1]{%
\externaldocument{#1}%
\addFileDependency{#1.tex}%
\addFileDependency{#1.aux}%
}
\myexternaldocument{main}
\myexternaldocument{supp_cr_arxiv}

\usepackage[breaklinks=true,bookmarks=false, pagebackref, colorlinks]{hyperref}

\usepackage[capitalize]{cleveref}
\crefname{section}{Sec}{Secs.}
\Crefname{section}{Section}{Sections}
\Crefname{table}{Table}{Tables}
\crefname{table}{Tab.}{Tabs.}

\iccvfinalcopy 

\def\iccvPaperID{7265} 

\ificcvfinal\fi

\begin{document}

\title{Learning to Learn: How to Continuously Teach Humans and Machines}

\author {
    Parantak Singh\textsuperscript{\rm 1, 2},
    You Li\textsuperscript{\rm 2,3},
    Ankur Sikarwar\textsuperscript{\rm 1, 2},
    Weixian Lei\textsuperscript{\rm 4},
    Difei Gao\textsuperscript{\rm 4},\\
    Morgan B. Talbot\textsuperscript{\rm 5, 6},
    Ying Sun\textsuperscript{\rm 2},
    Mike Zheng Shou\textsuperscript{\rm 4},
    Gabriel Kreiman\textsuperscript{\rm 5},
    Mengmi Zhang\textsuperscript{\rm 1, 2}\\
    \textsuperscript{\rm 1} \small Nanyang Technological University (NTU), Singapore
    \textsuperscript{\rm 2} \small CFAR and I2R, Agency for Science, Technology and Research, Singapore,\\
    \textsuperscript{\rm 3} \small University of Wisconsin-Madison, USA,
    \textsuperscript{\rm 4} \small Show Lab, National University of Singapore, Singapore,  \\ 
    \textsuperscript{\rm 5} \small Boston Children's Hospital, Harvard Medical School, USA,
    \textsuperscript{\rm 6} \small Harvard-MIT Health Sciences and Technology, MIT,\\
    \small Address correspondence to mengmi@i2r.a-star.edu.sg
}
\maketitle
\ificcvfinal\thispagestyle{empty}\fi

\begin{abstract}
Curriculum design is a fundamental component of education. For example, when we learn mathematics at school, we build upon our knowledge of addition to learn multiplication. These and other concepts must be mastered before our first algebra lesson, which also reinforces our addition and multiplication skills. Designing a curriculum for teaching either a human or a machine shares the underlying goal of maximizing knowledge transfer from earlier to later tasks, while also minimizing forgetting of learned tasks. Prior research on curriculum design for image classification focuses on the ordering of training examples during a single offline task. Here, we investigate the effect of the order in which multiple distinct tasks are learned in a sequence. We focus on the online class-incremental continual learning setting, where algorithms or humans must learn image classes one at a time during a single pass through a dataset. We find that curriculum consistently influences learning outcomes for humans and for multiple continual machine learning algorithms across several benchmark datasets. We introduce a novel-object recognition dataset for human curriculum learning experiments and observe that curricula that are effective for humans are highly correlated with those that are effective for machines. 
As an initial step towards automated curriculum design for online class-incremental learning, we propose a novel algorithm, dubbed Curriculum Designer (CD), that designs and ranks curricula based on inter-class feature similarities.  
We find significant overlap between curricula that are empirically highly effective and those that are highly ranked by our CD.  
Our study establishes a framework for further research on teaching humans and machines to learn continuously using optimized curricula.
Our code and data are available through \href{https://github.com/ZhangLab-DeepNeuroCogLab/Learning2Learn}{this link}. 
\end{abstract}


\begin{figure}
    \begin{center}
    \includegraphics[width=0.4\textwidth]{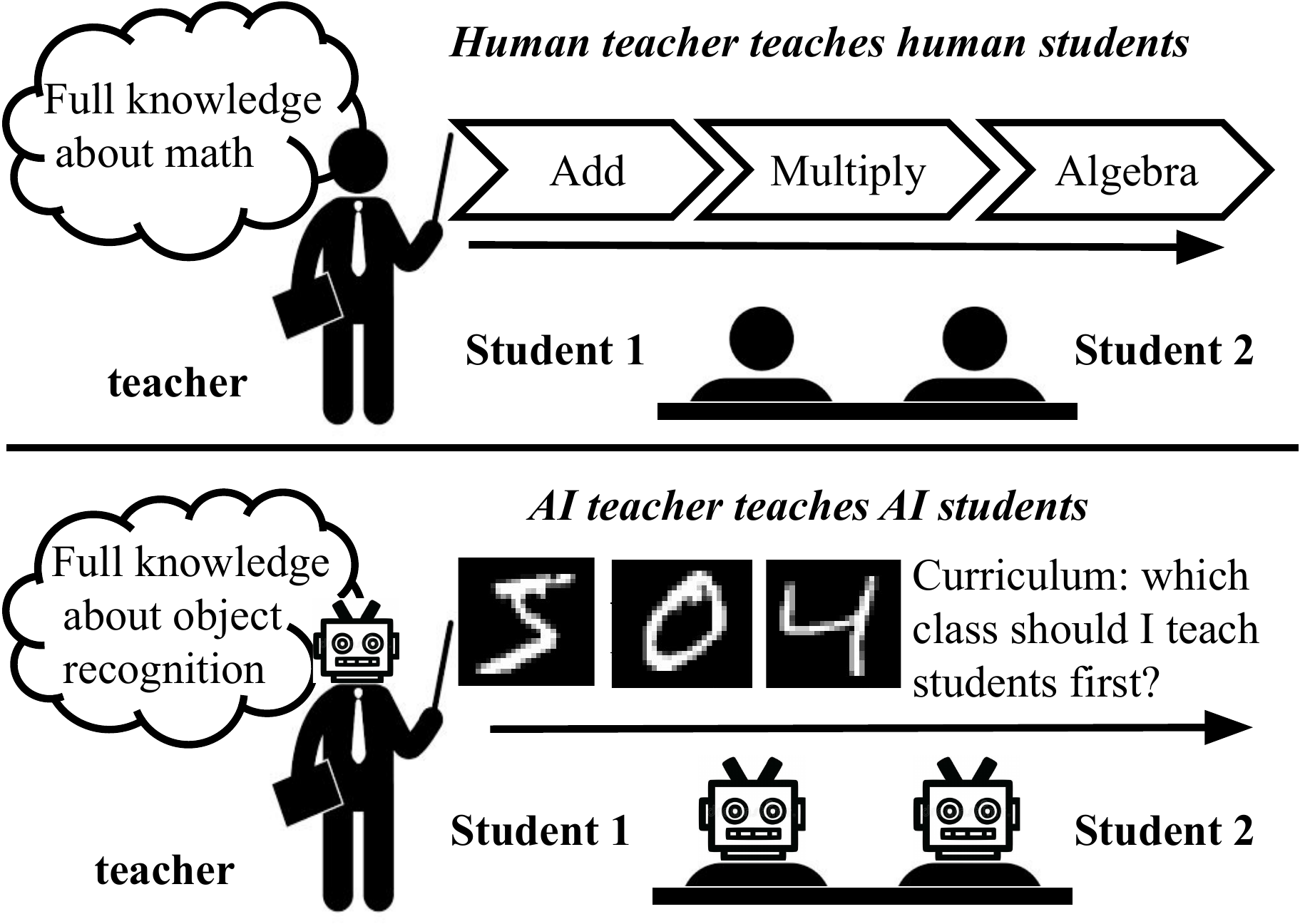}
    \caption{\textbf{Curricula in classroom and machine learning settings}. In human education, a natural curriculum designed by a knowledgeable math teacher prescribes teaching, in order, addition, multiplication, and algebra. Student 1 and Student 2 learn these concepts in a continuous fashion.     
    Similarly, in an image classification task, what is the optimal curriculum for an AI teacher to continuously teach AI students to recognize images? 
    }
    \label{fig:fig1}
    \end{center}
    \vspace{-10mm}
\end{figure}

\vspace{-5mm}
\section{Introduction}


\label{sec:intro}
When learning mathematics, students continuously advance through a curriculum that guides them to first learn addition, then multiplication, and later algebra such that each new concept both builds upon and reinforces existing knowledge (\textbf{Fig~\ref{fig:fig1}}). Studies on curriculum development in education show that careful design of curricula for human students can enable an incremental learning process, facilitating positive knowledge transfer to new tasks and minimizing forgetting of learned tasks \cite{siraj2002researching}. Drawing on this inspiration, our goal is to develop a knowledgeable artificially intelligent (AI) teacher (a ``curriculum designer") that produces optimized curricula that enhance learning outcomes of both human students and machine learning algorithms (``AI students'').

A growing body of literature in the field of ``curriculum learning'' investigates the order in which training examples are presented to machine learning (ML) algorithms. The effects of curriculum on ML outcomes have been explored in supervised~\cite{soviany2021curriculum, zhou2021robust, wang2022efficient, xiang2020learning, bell2022effect}, weakly-supervised~\cite{tudor2016hard, shu2019transferable, guo2018curriculumnet}, unsupervised~\cite{yang2020curriculum, soviany2021curriculum, sakaridis2019guided}, and
reinforcement learning (RL)~\cite{klink2020self, florensa2017reverse, qu2018curriculum} settings. Existing work in supervised learning~\cite{soviany2021curriculum, zhou2021robust, wang2022efficient, xiang2020learning, bell2022effect} has demonstrated improved generalization ability and convergence speed through the design of more effective curricula, but only by estimating intra-class example difficulty and scheduling examples within a \textit{single} task. Unlike supervised classification algorithms that require multiple passes over large, shuffled training datasets to learn many classes in parallel, humans learn a variety of tasks incrementally through a continuous stream of non-repeating experience. This process is more closely emulated in continual learning (CL) settings, where ML algorithms learn a series of tasks one at a time, and particularly in online CL settings where each training example is shown only once \cite{mai2022online}. Although the presentation order of separate tasks is a central focus in designing curricula for humans, the influence of task order on offline and online CL outcomes remains largely unexplored.

To address this question, we investigated the effects of class presentation order (``curriculum") during online class-incremental CL 
by machines and humans. An ideal learning algorithm in this setting would leverage its knowledge of early tasks to more effectively learn later tasks (forward transfer) while also avoiding forgetting early tasks. The challenging problem of ``catastrophic forgetting'' in artificial neural networks has been addressed with a variety of CL-specific algorithms \cite{wang2023comprehensive}. Since each CL algorithm modulates the learning process using a different strategy, we conceptualize different CL algorithms as distinct AI students that may or may not maximally benefit from the same curricula. 
Our empirical ML results suggest that curriculum design choices greatly influence knowledge transfer and forgetting across CL algorithms and hyperparameter settings of each. We demonstrate a strong correlation among different CL algorithms in the relative effectiveness of different curricula. We also found curriculum effects that are correlated among CL algorithms in a continual visual question answering setting ~\cite{lei2022symbolic}. 

Building upon these findings, we propose an automatic curriculum designer (CD), an algorithm that efficiently designs and ranks curricula. 
In a nutshell, our CD enables pairs of object classes that are nearer to each other in feature space to be separated farther from each other in time during the training processes of neural networks and humans. Unlike pre-defined curriculum learning algorithms~\cite{tudor2016hard, lotfian2019curriculum, wei2021learn, soviany2020image}, our CD does not require prior knowledge from domain experts, nor any human intervention. Our results demonstrate that curricula ranked highly by our CD improve learning performance across multiple CL algorithms. 


To probe further whether the optimal curricula for continual machine learning are also beneficial for human learning,
we conducted a series of human psychophysics experiments and contributed a new novel-object recognition CL benchmark. From the experiments,
we observed a high degree of agreement between the most effective curricula for CL algorithms and humans.  

Our main contributions to this work are as follows:
\begin{itemize}[noitemsep,nolistsep]
\item We establish a methodology to study curriculum effects in online class-incremental learning.
\item We introduce a new novel-object recognition dataset to benchmark the effectiveness of class-incremental curricula for humans and CL algorithms.
\item We quantify commonalities among empirically optimal curricula for CL algorithms and humans.
\item We propose an automated curriculum designer that can design the optimal curricula and rank (score) the existing curricula by their effectiveness. 
\end{itemize}


\section{Related Works}
\label{sec:relatedworks}

\subsection{Continual Learning (CL)}\label{sec:SLCL}

CL strategies can be grouped into three categories: weight regularization, replay, and architecture expansion. 
Regularization methods constrain or regularize weight updates during training on new tasks using information from previous tasks
~\cite{li2017learning,chaudhry2018efficient,he2018overcoming,kirkpatrick2017overcoming,zenke2017continual,lee2017overcoming}. 
Replay-based strategies involve storing a subset of examples from previous tasks and interspersing them with training data from newly encountered tasks to mitigate forgetting~\cite{wu2019large,rebuffi2017icarl,aljundi2019gradient, chaudhry2018efficient, nguyen2017variational,lopez2017gradient,bang2021rainbow}.
Architecture adaptation methods involve expanding or restructuring neural networks to assimilate new tasks~\cite{li2017learning,he2018overcoming,kirkpatrick2017overcoming,zenke2017continual,lee2017overcoming,golkar2019continual,schwarz2018progress,fernando2017pathnet, rajasegaran2019random, serra2018overcoming, adel2019continual}. 
CL methods are predominantly evaluated in \textit{offline} class-incremental settings where many passes over data within each task are permitted. Researchers report average performance over multiple runs with \textit{random} class orders. Here, we exhaustively study the effect of class presentation order during \textit{online} class-incremental learning, where only one pass over the data within each task is allowed. 

\subsection{Curriculum Learning}


Curriculum learning refers to learning with a meaningful ordering of training examples, commonly from ``easier" to ``harder" data \cite{bengio2009curriculum, allgower2012numerical}. The efficacy of proposed curricula is evaluated in terms of generalization to test data and convergence speed during training. 
Previous works in curriculum learning can be categorized into predefined curriculum learning \cite{bengio2009curriculum, soviany2020image, chen2015webly, choi2019pseudo} and automatic curriculum learning \cite{wang2021survey, kim2018screenernet, fan2018learning, graves2017automated}.
Predefined curriculum learning entails designing a data scheduler or a difficulty measure with human priors. These algorithms work well when designed for specific tasks, but generalize poorly to out-of-domain tasks. In contrast, we propose an automatic curriculum designer that can design and rank curricula based on inter-class feature differences.  

In automatic curriculum learning, most works adopt data-driven approaches \cite{kim2018screenernet, fan2018learning, graves2017automated} and RL-based approaches incorporating student feedback~\cite{singla2021reinforcement, he2021quizzing, doroudi2019s, mu2021automatic, sen2018teaching}. These methods are often deployed in teaching both machines
\cite{tudor2016hard, shu2019transferable, guo2018curriculumnet,yang2020curriculum, soviany2021curriculum, sakaridis2019guided,klink2020self, florensa2017reverse, qu2018curriculum} and humans \cite{singla2021reinforcement, he2021quizzing, doroudi2019s, mu2021automatic, sen2018teaching}. In image classification settings, curriculum learning approaches are almost exclusively oriented toward measuring intra-class example difficulty. Existing methods specifically focus on a single multi-class object recognition task \cite{wu2021when, tang2012self, saxena2019data, guo2018curriculumnet} in which all examples from each class can be trained on multiple times. We deviate from previous studies in examining the order in which classes or tasks are presented to the network, rather than the ordering of training examples within one task. 


One recent study highlighted how the most widely-used curriculum design strategy (increasing difficulty) may not always be optimal, and how anti-curricula (``harder" to ``easier") or random orderings yield comparable results in multi-class image classification settings \cite{wu2021when}. The study reported that curriculum effects become stronger when the number of training iterations is limited. Aligned with this constraint, we investigated the effect of curriculum on CL algorithms under stringent online conditions where training is limited to a single pass through the data. 

\section{Experiments}\label{sec:exps}

We conducted our experiments in the
online class-incremental 
learning setting. An image dataset $D$ comprises $N$ object classes $\{c_1,c_2\cdots c_N\}$ with $K$ training images each. The objective is to propose a temporal order of class presentation $T$ from $t_1,t_2\cdots t_N$ (a ``curriculum") such that a given CL algorithm $\mathcal{A}$ (a ``student") yields the optimal learning outcome. That is, $\mathcal{A}$ learns to adapt to new classes with minimal forgetting of previously learned classes while progressing through $T$. 

\begin{figure*}[t]
\begin{center}
\includegraphics[width=0.85\textwidth]{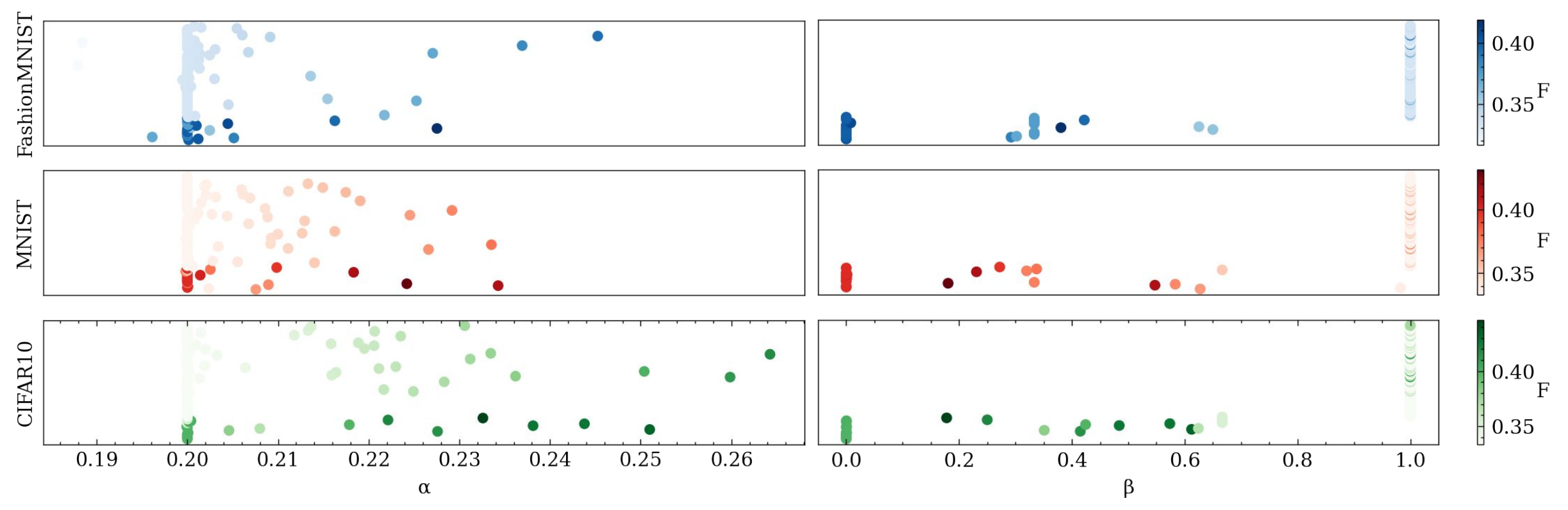}\vspace{-7mm}
\end{center}
  \caption{\textbf{Curricula influence the learning efficacy of the Vanilla CL algorithm (Sec~\ref{sec:algos}) across MNIST, FashionMNIST, and CIFAR10 datasets (Sec~\ref{sec:datasets})}. We trained the vanilla CL algorithm on all curricula from each dataset. Each dot represents one curriculum. We report the distribution of average accuracy $\alpha$ over all the seen classes
  \textbf{(left panel, Sec~\ref{sec:eval})}
  and
  the distribution of forgetfulness $\beta$ at the last task
  \textbf{(right panel, Sec~\ref{sec:eval})}. We introduced $\mathcal{F}$ as the measure of the learning efficacy of a given curriculum (\textbf{Sec~\ref{sec:eval}}). See the colorbar on the right for different $\mathcal{F}$ values. 
  Note that the y-axis does not carry any meaning. All the dots are randomly spread along the y-axis for easy visualization of the $\alpha$ and $\beta$ distributions. 
  }\vspace{-4mm} 
\label{fig:fig2}
\end{figure*}

\subsection{Datasets and Baselines} \label{sec:datasets}
We used three datasets for our experiments: MNIST ($60,000$ training images, $10,000$ test images) \cite{lecun1998mnist}, FashionMNIST ($60,000$ training and $10,000$ test images) \cite{xiao2017fashion}, and CIFAR10 ($50,000$ training and $10,000$ test images) \cite{krizhevsky2009learning}. 
Each dataset consists of 10 object classes. Ideally, each curriculum is a permutation of 10 object classes, resulting in a total of 10! (more than $3e^{6}$) possible curricula per dataset. Thus, running all permutations is infeasible due to limited computational resources. To mitigate this issue, we introduced two paradigms: in ``paradigm-I'', we chose a subset of the dataset comprising 5 classes with 1 class per task, and in ``paradigm-2'', we made 5 tasks with 2 classes each. In both paradigms, the order of the exemplars from the classes within a task is fixed and only the task sequence is permuted, resulting in a total of $5! = 120$ curricula.
Without loss of generality, we only present and discuss results for paradigm-I. See 
\textbf{Sec~\ref{sec:datasets_supp}} for details of class grouping, and 
see \textbf{Sec~\ref{sec:optimal_curricula}}-\textbf{\ref{sec:cd.design}}, and \textbf{Fig~\ref{fig:fig.top5.mnist.10}}-\textbf{\ref{fig:fig.top5.cifar.10}}, \textbf{\ref{fig:fig.line.mnist.10}}-\textbf{\ref{fig:fig.curricula.agreement}}, \textbf{\ref{fig:fig.top10_bot10_paradigmII}}, \textbf{\ref{fig:fig.recall@allk_10}}, \textbf{\ref{fig:fig.recall@k}} for results in paradigm-I. In general, the conclusions drawn in the first paradigm also hold true in the second. In paradigm-I, we used classes `$0$,' `$1$,' `$2$,' `$3$,' and `$4$' from MNIST, classes `coat,' `dress,' `pullover,' `top,' and `trouser' from FashionMNIST, and classes `airplane,' `automobile,' `bird,' `cat,' and `deer' from CIFAR10. 

As we are the first to study curriculum learning in online class-incremental learning, we used a random curriculum designer 
as our baseline. The random designer randomly ranks the 120 curricula for each dataset. We repeated the random designers over 100 times with different random seeds, resulting in 100 sets of 120 randomly ranked curricula per dataset.

\subsection{Continual Learning Algorithms} \label{sec:algos}
Among the CL algorithms surveyed in \textbf{Sec~\ref{sec:SLCL}}, we chose two weight regularization methods: 
Elastic Weight Consolidation (EWC) \cite{kirkpatrick2017overcoming} and Learning without Forgetting (LwF) \cite{li2017learning}. 
EWC estimates the importance of all weights after each task and penalizes weight updates in proportion to their prior importance in the loss function. LwF uses the knowledge distillation loss~\cite{hinton2015distilling} to regularize the current loss with soft targets acquired from a preceding version of the model. Replay-based CL algorithms involve joint training on old and new samples and often yield superior performance. We thus also include one replay method, where the images from previous tasks are randomly selected for the memory buffer and intermixed with the training data in the current task for replays. We fix the memory buffer size constant over all the tasks, which approximately equals the size of storing 2\% of the entire training set in each dataset. 
See curriculum analysis of the replay method in \textbf{Sec~\ref{sec:replay}} and \textbf{Fig~\ref{fig:fig.replay}}. 
However, these results should be interpreted with caution since the replay sequence of replay data interferes with the fixed class order in a given curriculum. We evaluate EWC, LwF, and naive replay alongside a ``vanilla'' fine-tuned method without any measures to prevent catastrophic forgetting.

The objective of this paper is not to exhaustively compare the performance of CL algorithms, but to study how curriculum affects the learning mechanism of each algorithm. 
For fair comparisons, we used a frozen SqueezeNet \cite{iandola2016squeezenet} pre-trained on a subset of 100 classes from ImageNet \cite{deng2009imagenet} (ImageNet100) as the feature extractor for all three CL algorithms. We ensured that the 100 classes used for pre-training do not overlap with any of the classes selected for our CL experiments (\textbf{Sec~\ref{sec:datasets}}). The fine-tunable classification layers for all CL algorithms were initialized with the same set of random weights prior to continual training. 
Results in \textbf{Sec~\ref{sec:results}} are reported based on the performance of the three selected CL algorithms over 3 independent runs with different random seeds. 

We used the standard public implementations of each CL algorithm from 
\cite{lomonaco2021avalanche}. 
Note that the online CL results reported in our paper deviate from the original CL results in \cite{lomonaco2021avalanche}, because each training example can be seen only once in the online setting. All three CL algorithms are trained using the Adam optimizer with a learning rate of $1e^{-3}$. We performed hyperparameter searches for all CL algorithms. See \textbf{Sec~\ref{sec:results.4}} for results and discussions about hyper-parameter variations. However, we emphasize that each CL algorithm with a different set of hyper-parameters is conceptualized as a different ``student." Though the same curriculum can be applied to all CL algorithms, the learning outcomes for different students might vary. 

\subsection{Evaluation Metrics} \label{sec:eval}


\noindent \textbf{Learning Effectiveness $\mathcal{F}$.}
An effective CL algorithm quickly adapts to new classes with minimal forgetting of previously learned classes. To evaluate the learning efficacy of a CL algorithm for a given curriculum, we introduced the effectiveness score $\mathcal{F}$. The metric $\mathcal{F}$ accounts for two aspects: (1) the average accuracy $\alpha$ over all seen classes should be as high as possible, and (2) the accuracy difference $\beta$ on the test images from the first task between the first task and the last task should be as small as possible. We formulate $\mathcal{F}$ as $\frac{2}{\beta + \frac{1}{\alpha}}$. $\mathcal{F}$ considers contributions from both $\alpha$ and $\beta$, while penalizing extreme values. 


We report the distribution of $\mathcal{F}$ for all curricula over three datasets in \textbf{Fig~\ref{fig:fig2} and Sec~\ref{sec:results.1}}. We see that a curriculum with high $\mathcal{F}$ (darker dots) has high $\alpha$ (\textbf{Fig~\ref{fig:fig2}}, left panel)  and low $\beta$ (\textbf{Fig~\ref{fig:fig2}}, right panel), highlighting how $\mathcal{F}$ reflects the overall learning effectiveness of a CL algorithm. We also reported $\mathcal{F}$ as a function of number of tasks (\textbf{Sec~\ref{sec:f_overtime}} and \textbf{Fig~\ref{fig:fig.f_overtime}}) and found that the curriculum effect becomes more prominent with longer task sequences. 




\noindent \textbf{Recall@K.} 
We used Recall@K to assess the teaching effectiveness of our curriculum designer (CD, \textbf{Sec~\ref{sec:model}}). 
Recall@K calculates the proportion of overlap between the top-K recommended curricula by our CD among the union set of all the top-K empirically ranked curricula by all $\mathcal{A}$s. We used the empirical curriculum rankings of EWC, LwF, and Vanilla for these calculations. 
Recall@K ranges from 0 to 1, where a higher value indicates better CD performance. Note that Recall@K also depends on the similarity of the curriculum effect among different CL algorithms. 

Recall@K quantifies our CD's ability to identify the top-k empirically ranked curricula, but is not influenced at all by the rankings of less effective curricula. We argue that the CD's rank order among the most effective curricula is of special importance, particularly for applications where the goal is simply for the CD to find the most effective possible curriculum. We nonetheless include supplementary results for Spearman's rank correlation coefficient, which assesses the degree of agreement in rankings across all curricula (see \textbf{Sec~\ref{sec:spearman}}). One disadvantage of both Recall@K and rank correlation coefficients is that they do not account for the similarities between the curricula themselves. In the next section, we introduce the discrepancy measure $\mathcal{H}$ as a complementary measure that addresses this issue. 


\noindent \textbf{Curriculum Discrepancy $\mathcal{H}$.} \label{sec:curriculum_discrepancy}
To assess the consistency between two sets of ranked curricula, we propose the curriculum discrepancy measure ($\mathcal{H}$), inspired by gene sequence comparison methods \cite{bonham2014alignment}. $\mathcal{H}$ quantifies the dissimilarity between two sets of ranked curricula. Curriculum rankings are either determined by a CD or empirically determined based on $\mathcal{F}$ after exhaustively running $\mathcal{A}$ on all curricula of a given dataset.
 

We sort curricula using $\mathcal{F}$ in ascending order, and divide the range of $\mathcal{F}$ into 5 uniformly-sized bins or ``tiers.'' 
Since studying the characteristics of the most effective curricula is critically important for the benefits of human and machine learning, in this work we focus on analyzing the curriculum discrepancy $\mathcal{H}$ from the top tier with the highest $\mathcal{F}$. 

To calculate $\mathcal{H}$, we first assign each object class to a unique letter identifier and convert each curriculum to a string. As an example, $5$ object classes in a dataset can be represented with letters $A,\ B,\ C,\ D,\ \text{and } E$. Any curriculum can then be represented as a combination of these $5$ letters, such as $ABCED$ for curriculum 1 and $DECBA$ for curriculum 2. For a ranked curriculum set in the top tier, we can 
concatenate all the curricula into one string. In the example above, we have $ADBECCEBDA$. Given a pair of strings (two sets of ranked curricula), we use the Hamming distance to measure their curriculum discrepancy $\mathcal{H}$. The lower the $\mathcal{H}$ value, the higher the consistency: if the two ranked curricula are in exactly the same order, $\mathcal{H} = 0$. Note that Recall@K and ranking metrics like NDCG \cite{jarvelin2002cumulated} and rank correlations \cite{zar2005spearman} focus solely on comparing the order in which curricula are ranked, without reference to similarities among class orderings within curricula. 
We are unaware of any existing metrics that address rank similarities both within and between curricula. 


In \textbf{Fig~\ref{fig:fig2}}, we observe a skewed distribution of $\mathcal{F}$ where there are a few curricula with very high $\mathcal{F}$ but many curricula with similarly low $\mathcal{F}$s. Thus, different tiers have different numbers of curricula. For a pair of ranked curricula sets in tier 5 where each set may have a different number of curricula, we choose the number of curricula in one set as a reference and compare it with the other curricula set containing an equal number of curricula. We do this once with each of the sets as the reference. The mean is then reported as the $\mathcal{H}$ for this pair of ranked curricula sets.

We conducted statistical tests for all experiments involving the above evaluation metrics, and report the results in \textbf{Sec~\ref{sec:stat_anal}}.


\begin{figure*}[t]
\centering
\subfloat[Example objects]{\includegraphics[width= 4.4cm]{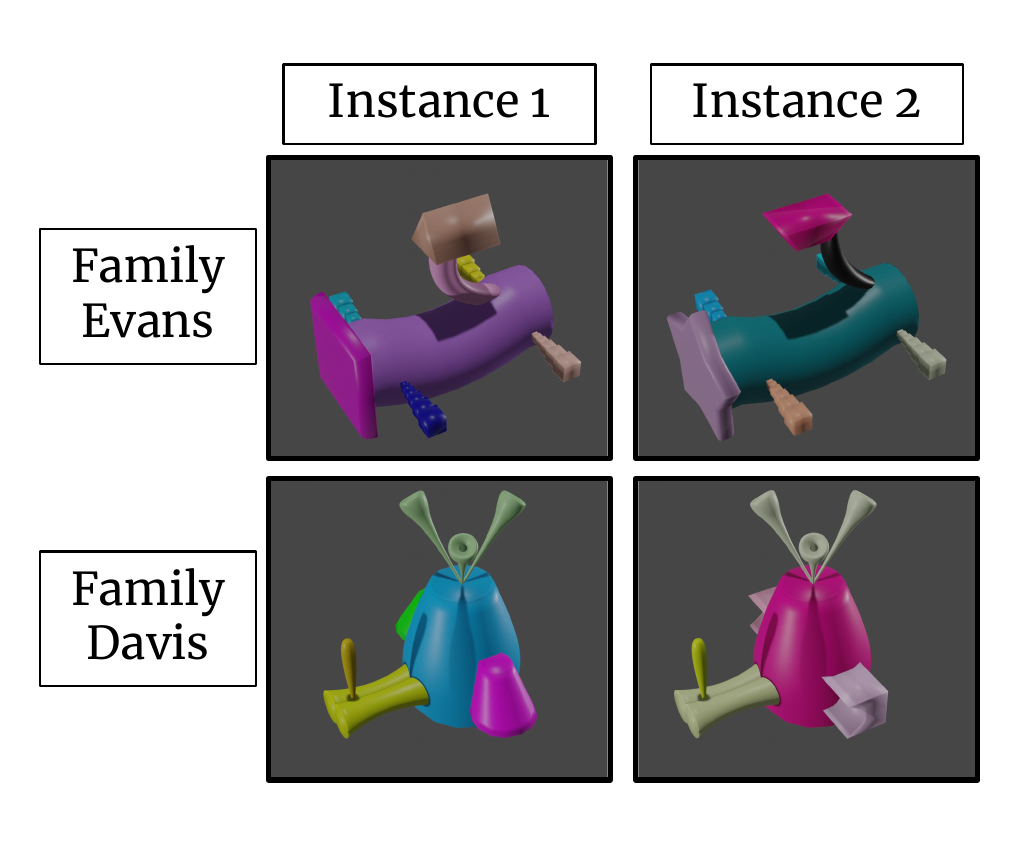}\label{fig:fig6a}}\hspace{0.2cm}
\subfloat[Class incremental learning setting]{\includegraphics[width= 6cm]{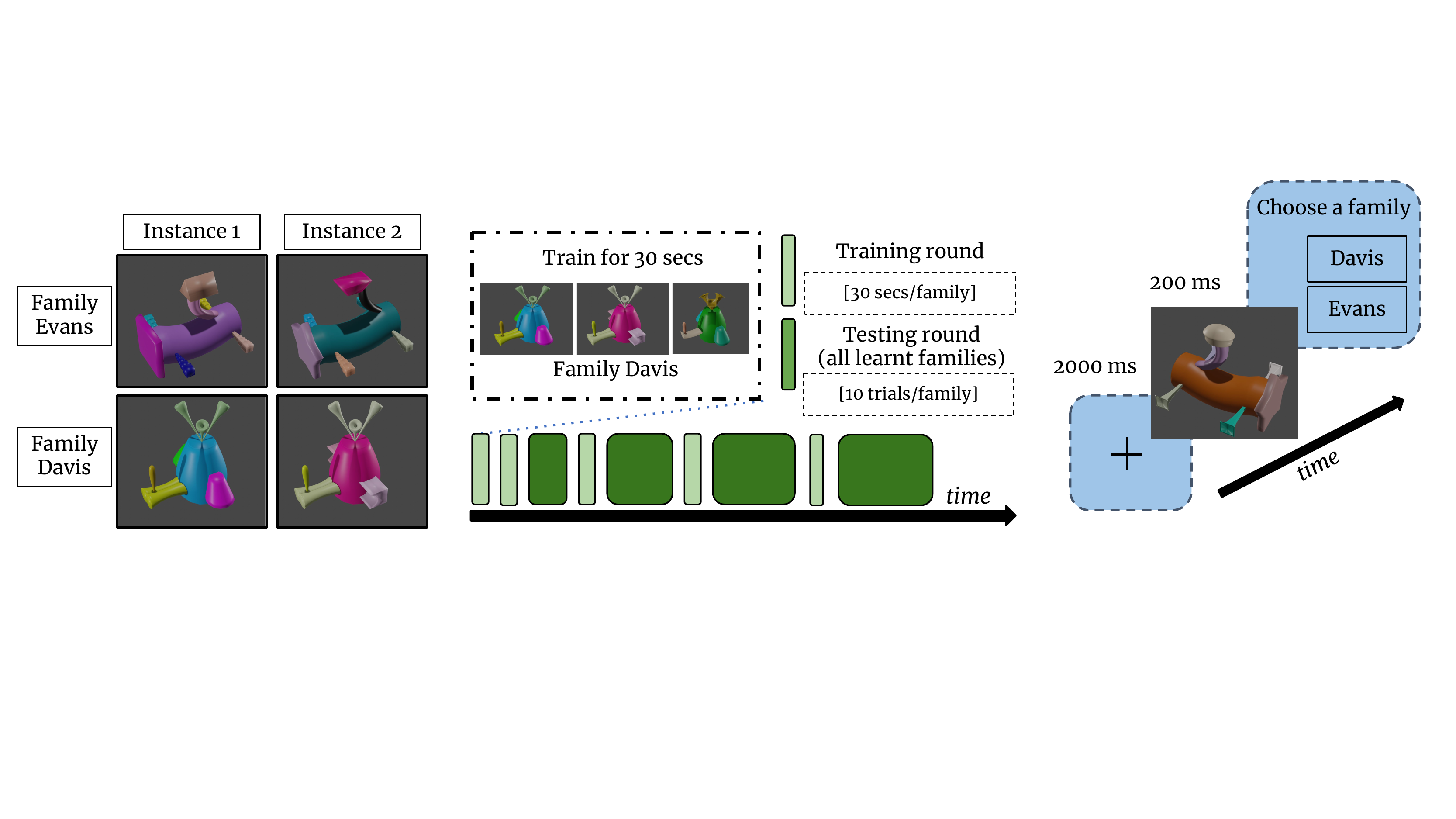}\label{fig:fig6b}}\hspace{0.2cm}
\subfloat[Test trial schematics]{\includegraphics[width= 4cm]{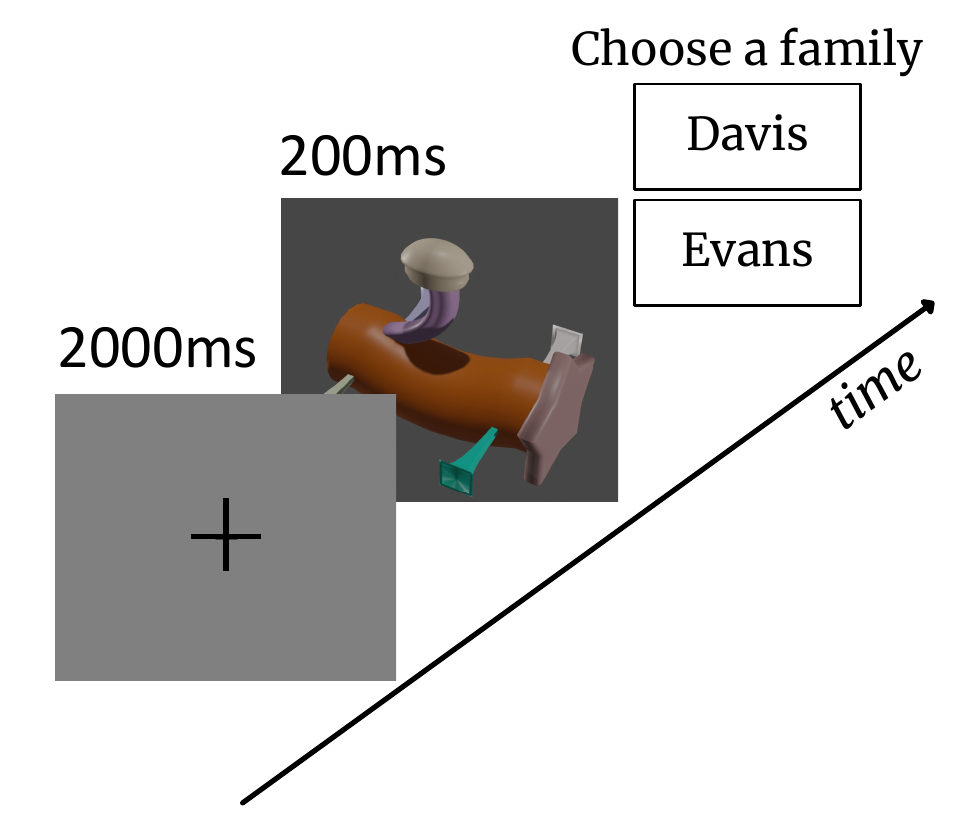}\label{fig:fig6c}}\vspace{-0.2cm}
   \caption{\textbf{Overview of human behavioral experiments in a class incremental setting}. (a) Two example object instances from each of two families in the Novel Object Dataset (NOD, \textbf{Sec~\ref{sec:human_bench}}). (b) Experiment schematic. Subjects progressed through 4 tasks, each with a training and testing round. During training, subjects were presented with three rotating object instances per family for 30 seconds, with the goal of being able to recognize the objects presented in the testing round. 
   In the first training round, 2 families were introduced. In subsequent training rounds, one additional family was introduced per task, without showing instances from previously learned families. During testing, subjects were tested on 10 trials from each learned family. The trial order was randomly shuffled during testing. (c) In each test trial, subjects were presented with a fixation cross (2000ms) followed by the stimulus (200ms). After the image offset, subjects were asked to choose the family of the presented object among all previously encountered families. 
   }
\label{fig:fig6}\vspace{-4mm}
\end{figure*}


\subsection{Human Benchmark} \label{sec:human_bench}

\noindent \textbf{Novel Object Dataset (NOD)}

We introduce the Novel Object Dataset (NOD) containing novel 3D objects with a categorical structure to test the continual learning abilities of humans and continual learning algorithms. 
NOD is a subset of the larger ``Fribbles" dataset \cite{barry2014meet}. The dataset comprises 5 object families with 5 object instances per family. The instances and families differ in their main body structure and in the locations and shapes of various appendages (\textbf{Fig~\ref{fig:fig6a}}).
We used Blender \cite{filippov2018blender} 
to load the 3D object meshes, and rendered a 1920 $\times$ 1080 sized image of each object for every 10 degrees of azimuth and every 10 degrees of elevation, resulting in a total of 32,400 images ($36^2$ images per instance). We rendered the objects against a grey background to avoid confounding factors such as background biases. We randomly colored every object instance's body and appendages separately. To make the families easier for subjects to remember, we assigned a commonly used surname to each family.


\noindent \textbf{Psychophysics Experiments}

Following standard protocols approved by our Institutional Review Board, we evaluated human performance on NOD using Amazon Mechanical Turk (MTurk) with the subjects' informed consent. 
The experiment duration on average was 20 minutes. Each participant was compensated. For quality control purposes, we also conducted in-lab experiments. We report the results from MTurk here and provide the details and results of the in-lab experiments in \textbf{Sec~\ref{sec:hum_benchmark}} and \textbf{Fig~\ref{fig:amt_interface}}-\textbf{\ref{fig:fig.line.novelnet}}, \textbf{\ref{fig:fig.top_bot.nod.mturk_inlab}}, \textbf{\ref{fig:fig.topbot.mturk.inlab}}. The in-lab results support the conclusions drawn from the MTurk experiments.

We divided the experiment into 4 tasks, such that the first task had 2 object families and each subsequent task had 1 object family; this makes a total of $\binom{5}{2} \times 3! = 60$ possible curricula. Each subject is randomly assigned a curriculum. We recruited 242 subjects for a total of 34,848 test trials, with an average of 4 subjects tested on each curriculum. A schematic of the experiment is illustrated in \textbf{Fig~\ref{fig:fig6b}}. During the training rounds, the subjects were presented with 3 object instances per family
that were shown rotating continuously along the azimuth. During the testing rounds, the subjects were shown a 640 $\times$ 480 sized GIF for each trial from the remaining 2 object instances per family (\textbf{Fig~\ref{fig:fig6c}}). Train and test instances differ.
We took several precautions to ensure data quality and that subjects paid attention to the experiments (see \textbf{Sec~\ref{sec:hum_benchmark}}). 
Despite our simple stimulus design, we found that the majority of the participants ranked the experiments as difficult with an average difficulty score of 6.8/10 (10 = max. difficulty).

\section{Curriculum Designer}\label{sec:model}

We propose a proof-of-concept model, a Curriculum Designer (CD) for online class-incremental learning. 
Given a curriculum, our CD assigns a ranking score based on inter-class feature similarity. Our CD scores all possible curricula to produce a ranked set of curricula for each dataset.  
The low discrepancy in the ranked curricula of different continual learning algorithms (see the results in \textbf{Sec~\ref{sec:results.4}}) suggests that our CD does not necessarily need to depend on the feedback of a specific learning algorithm $\mathcal{A}$. The objective of our CD is to propose a universal curriculum that improves learning outcomes of any given $\mathcal{A}$ relative to the average of randomly chosen curricula. 

\subsection{Feature Distance Confusion Matrix} \label{sec:model.1}

Given an curriculum defined as $c_{t=1},c_{t=2},...,c_{t=N}$,
our CD uses an inter-class distance confusion matrix $M$ of size $N\times N$, where any 
element $M_{(i,j)}$ represents a distance measure between two class prototypes, $c_{t=i}$ and $c_{t=j}$. To calculate a class prototype vector for each class, we used a teacher network to extract features from all images of the given class and took the vector mean. 
The feature distance $M_{(i,j)}$ between each pair of class prototypes $c_{t=i}$ and $c_{t=j}$ is calculated with the cosine distance.
We conducted ablation experiments on distance metrics (\textbf{Sec~\ref{sec:results.3}}). In practice, extracting features from all images in a large dataset 
is computationally costly. Thus, we randomly sampled 500 images per class to compute the prototypes. 


We used layers 1-12 of 2D-CNN SqueezeNet as our teacher network for computing class prototypes~\cite{iandola2016squeezenet}. Drawing on the analogy that a human teacher has full knowledge of the subject they teach, the teacher network is pre-trained on ImageNet \cite{deng2009imagenet}. For consistency with the learning algorithms themselves (\textbf{Sec~\ref{sec:algos}}), we fine-tuned the teacher network on the same set of 100 classes from ImageNet. The extracted feature vector of an input image is of size 1000. Prior knowledge of either the teacher or the student influences learning outcomes. We investigated the effect of prior knowledge
in \textbf{Sec~\ref{sec:results.3}}. 


\subsection{Ranking Curricula} \label{sec:model.2}
Given the inter-class distance confusion matrix $M$, we introduce a ranking score $s$ that keeps track of the accumulative advantage $v_{t}$ of choosing class $c_{t}$ at incremental step $t$ up to the final incremental step $N$: $s = \sum_{t=1}^{t=N} v_{t}$. Among all the curricula, the curriculum with the highest $s$ is selected as the optimal. Next, we introduce the design of the advantage $v_{t}$ for $c_{t}$ and its motivations.

Drawing on the idea of metric learning~\cite{chen2022multi} as well as the theoretical and practical foundations behind the impact of task ordering~\cite{lin2023theory, lee2021continual}, we choose the class $c_{t=1}$ at the first incremental step with the following criteria: the variance of the distances between the selected class prototype and the other classes' prototypes should be as small as possible.
Intuitively, lower class distance variance implies relatively similar distances to other classes: the first class is near the center of the multivariate class feature distribution. Starting to learn from the class comprising features shared with most other classes facilitates positive knowledge transfer when learning other classes at later steps. Thus, to encourage our CD to prioritize selecting the first class with the smallest distance variance, we define the advantage $v_{t=1}$ at the first incremental step as $1 - Var(\{M_{(1,j)}\}_{j=2}^{N})$, where $j$ is the corresponding class $c_j$ at incremental step $t=j$ 
and $Var(\cdot)$ is a function computing the variance from a set of distances. 

Subsequently, to eliminate catastrophic forgetting over incremental steps, we draw ideas from replay mechanism in CL \cite{wu2019large,rebuffi2017icarl,aljundi2019gradient, chaudhry2018efficient, nguyen2017variational,lopez2017gradient,bang2021rainbow} and select the last class $c_{t=N}$ based on the following criteria: the prototype of the selected class should have the smallest distance to $c_{t=1}$. The design motivation is to ensure that $c_{t=N}$ is the most similar to $c_{t=1}$ in terms of features. While $\mathcal{A}$ learns to classify $c_{t=N}$, these common features are functionally analogous to a feature replay of $c_{t=1}$, which regularizes the parameters of $\mathcal{A}$ to prevent forgetting.  Correspondingly, to encourage CD to prioritize replay-like class selection at the last incremental step, we define the advantage $v_{t=N}$ as  $1-M_{(N,1)}$.

Conversely, for the selection of the second class to learn at step $t=2$, we encourage CD to select the class whose prototype is the farthest away from its previous class $c_{t=1}$. This is in accordance with the classical notion in the curriculum learning literature that a curriculum should always be arranged in order, from easiest to the hardest~\cite{bengio2009curriculum}. The farther away the distance between two class prototypes, the easier it is for the algorithm $\mathcal{A}$ to learn the classification boundary between these two visually distinct classes. In this case, we define the advantage $v_{t=2}$ as $M_{(2,1)}$

We complete the ranking process of a given curriculum by iteratively performing the advantage evaluation back and forth over all subsequent incremental steps until we have examined all the classes. We summarize the piece-wise advantage function below:
\begin{equation*}
    v_t =
    \begin{cases}
        \mathsmaller{1 - \operatorname{Var}(\{M_{(1,j)}\}_{j=2}^{N})} & \text{, } t = 1 \\
        \mathsmaller{M_{t,t-1}} & \text{, } 1 < t \leq \lfloor \frac{N}{2} \rfloor \\
        \mathsmaller{1 - M_{t,N-t+1}} & \text{, } \lfloor \frac{N}{2} \rfloor < t \leq N
    \end{cases}
\end{equation*}
For every curriculum from a dataset, we compute its corresponding ranking score $s$ by summing the advantage for each class in a curriculum. Although it is daunting to perform heuristic searches for optimal curricula by exhaustively going through all possible curricula for a dataset, it is still computationally efficient for our CD given that it only scores curricula based on a 2D distance confusion matrix $M$.
See 
\textbf{Algorithm \ref{algo:your-algo}} (\textbf{Supp.}) for the pseudo-code of CD implementation.

\section{Results}\label{sec:results}

\subsection{Curriculum Strongly Impacts Performance} 
\label{sec:results.1}

\textbf{Fig~\ref{fig:fig2}} highlights the effect of curricula on the vanilla $\mathcal{A}$ (\textbf{Sec~\ref{sec:algos}}) over all three datasets (\textbf{Sec~\ref{sec:datasets}}). 
We observed a large variance in average accuracy $\alpha$, which ranged from $19\%$ to $26\%$ depending on the curriculum. This implies that curriculum strongly influences the overall performance over all tasks for the vanilla $\mathcal{A}$ (\textbf{Sec~\ref{sec:eval}}). 
$\beta$ reflects the degree of forgetting of the first task while learning later tasks (\textbf{Sec~\ref{sec:eval}}). The large variance in $\beta$ indicates that curriculum plays a significant role in preventing the vanilla $\mathcal{A}$ from forgetting the first class. The empirically optimal curriculum results in a more gradual decline in the accuracy on images from the initial task as subsequent tasks are introduced, which leads to a smaller $\beta$.

We introduced the learning effectiveness score $\mathcal{F}$, which incorporates both $\alpha$ and $\beta$ (\textbf{Sec~\ref{sec:eval}}). Darker dots in \textbf{Fig~\ref{fig:fig2}} indicate higher $\mathcal{F}$, generally implying larger $\alpha$ and smaller $\beta$. 
For example, 
for a model which learns the 1st task perfectly well and achieves 100\% accuracy but fails to adapt to any new tasks (0\% for the other four classes), we can calculate its effectiveness scores as:
$\alpha = (100\%+4 \times 0\%)/5 = 20\%$, $\beta = 100\% - 100\% = 0\%$ and $\mathcal{F} = 2/(0+5) = 0.4$. 
Another instance would be $\alpha =0.25$ but higher $\beta$, where the CL model learns a bit of each task and tends to forget previous tasks. The $\mathcal{F}$ differs by 0.09, 0.07 and 0.07  from the best to the worst curriculum for MNIST, FashionMNIST and CIFAR10.
These results from regularization-based CL algorithms $\mathcal{A}$s (\textbf{Sec~\ref{sec:algos}}) are constrained by the online class-incremental setting. Their $\mathcal{F}$ scores are in contrast to those of the highly effective replay method (\textbf{Sec~\ref{sec:algos}}) with an average $\mathcal{F} = 0.99, 0.87, 0.69$ on MNIST, FashionMNIST and CIFAR10, which often serve as upper bounds of continual learning performances. 
We present the distributions of $\alpha$, $\beta$, and $\mathcal{F}$ for EWC~\cite{kirkpatrick2017overcoming} and LwF~\cite{li2017learning} in \textbf{Sec~\ref{sec:curriculum_impact}} and \textbf{Fig~\ref{fig:fig.line.mnist.5}}-\textbf{\ref{fig:fig.f.5}}. The curricula trends observed in the discussion here are also applicable to these two algorithms.


\subsection{Our CD Predicts Optimal Curricula}
\label{sec:results.2}
\begin{figure}[t]
\centering
\includegraphics[width=0.4\textwidth, scale=1]{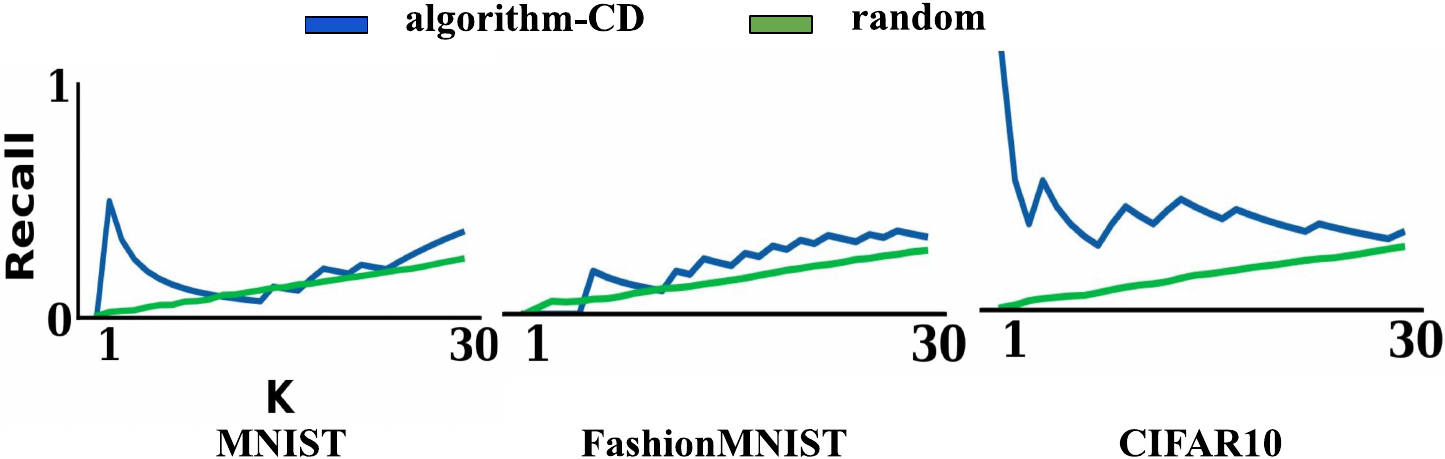}\vspace{-2mm}
  \caption{\textbf{Our Curriculum Designer (CD) predicts optimal curricula better than a random CD.} Recall@K (\textbf{Sec~\ref{sec:eval}}) of our CD (blue, \textbf{Sec~\ref{sec:model}}) and a random curricula designer (green) are reported as a function of K ranging from 1 to 30 across all three datasets (\textbf{Sec~\ref{sec:datasets}}), where K is the number of top curricula included in the metric. 
  }
\label{fig:fig3}\vspace{-4mm}
\end{figure}

To evaluate the effectiveness of the predicted curricula by our CD for CL algorithms $\mathcal{A}$s, we report results in terms of Recall@K (\textbf{Sec~\ref{sec:eval}}) in \textbf{Fig~\ref{fig:fig3}}.
We used a random curriculum designer
as a baseline for comparison to our CD. 
Across all three datasets, our CD (blue) outperformed the random model (green), particularly at small k values. Our CD achieves peaks in Recall@K of 0.5, 0.2, and 1 at K=2, K=5 and K=10 for MNIST, FashionMNIST and CIFAR10 respectively. 

Our results suggest that the CD performance does not depend on data complexity, as CD performs well on both MNIST and CIFAR10 despite CIFAR10 having more complex image features. Our curriculum designer exhibits remarkable performance on CIFAR-10. A plausible conjecture could be that these results are attributed to the striking resemblance between CIFAR-10 and ImageNet. The latter was employed for pre-training and served as the fundamental feature extractor for our curriculum designer.
We provide visualizations of the top-5 empirically-determined and CD-predicted curricula 
for all datasets in \textbf{Fig~\ref{fig:fig.top5.mnist}}-\textbf{\ref{fig:fig.top5.cifar.10}}.
The top curricula seem to align with the intuitions behind our CD design (\textbf{Sec~\ref{sec:model}}). Although our CD is effective in most cases, there is considerable room for improvement. We note that our CD has relatively weak performance on FashionMNIST, with Recall@K below the random CD for $K < 4$ and only slightly above random for $K \geq 4$.

\subsection{Analysis of CD Design Decisions}
\label{sec:results.3}

\begin{figure}[t]
\centering
\includegraphics[width=0.4\textwidth, scale=1]{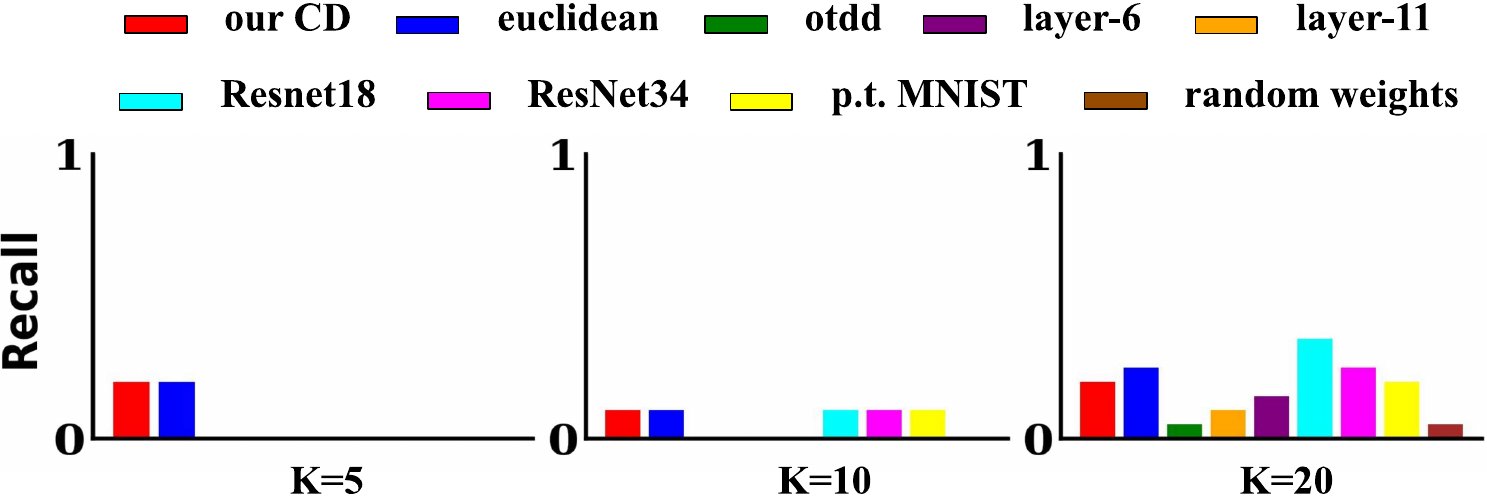}\vspace{-2mm}
  \caption{\textbf{Ablation results on our CD}. Recall@K bar plots for k=5, 10, and 20 with our CD and its ablations compared against the empirical curricula ranking determined by all continual learning algorithms $\mathcal{A}$s (\textbf{Sec~\ref{sec:algos}}) on MNIST (\textbf{Sec~\ref{sec:datasets}}) for paradigm-I (5 classes, \textbf{Sec~\ref{sec:datasets}}). 
  See \textbf{Sec~\ref{sec:results.3}} for the description of ablated CDs. 
  }
\label{fig:fig4}\vspace{-4mm}
\end{figure}

To evaluate the impact of individual design choices in our CD, we conducted experiments with variations of our CD on MNIST and presented the Recall@K results for K=5, 10, and 20 in \textbf{Fig~\ref{fig:fig4}}. 
First, instead of the cosine distance metric used in our CD, we changed the distance metric to Euclidean and Optimal Transport Dataset Distance (OTDD)~\cite{alvarez2020geometric} (euclidean and otdd).  
The ablated model with Euclidean outperforms OTDD and performs competitively well as our CD with cosine distance. This implies that the choice of measure for the inter-class distance is essential for curriculum designs. 
Next, we evaluated the effect of changing the layers used in the feature extractor to compute the distance confusion matrix $M$ by using layers 6 and 11 (layer-6 and layer-11). We observed that using layer-11 or layer-6, on average, leads to a performance decrement in recall at earlier Ks. This implies that the higher layers of the network produce more class-representative features that are useful for curricula ranking.
Furthermore, we replaced our default feature extractor SqueezeNet with ResNet34 and ResNet18~\cite{he2016deep}. Though the recall of these ablated models is not as high as our CD at K=5, they achieve a high recall at K=10. This implies that a change in architecture does not lead to dramatic performance deterioration in continual learning.

To study the effect of prior knowledge of our CD as the teacher,
 we introduce two variations.
First, we pre-trained the feature extractor of our CD on MNIST (p.t. MNIST). Compared with our original CD pre-trained on 100 classes of ImageNet (\textbf{Sec~\ref{sec:model.1}}), we did not observe any increase in recall at K=5; but we observed the high recall at K=10.
It is possible that the 100 classes from ImageNet share similar features with the classes from MNIST. Drawing on an example in pedagogy that a teacher with general math knowledge can teach arithmetic as efficiently as a teacher with only arithmetic-specific expertise, this experiment indicates that a teacher with broad knowledge in the field is as good as a teacher with area-specific knowledge. 
Next, we evaluate our CD with the weights of its feature extractor randomly initialized  (random-teacher). With the observation of the drastic drop in recall even at K=20, we conclude that prior knowledge of a teacher is indeed important for designing efficient curricula. 




\begin{figure}[t]
\begin{center}
\includegraphics[width=0.45\textwidth, scale=1]{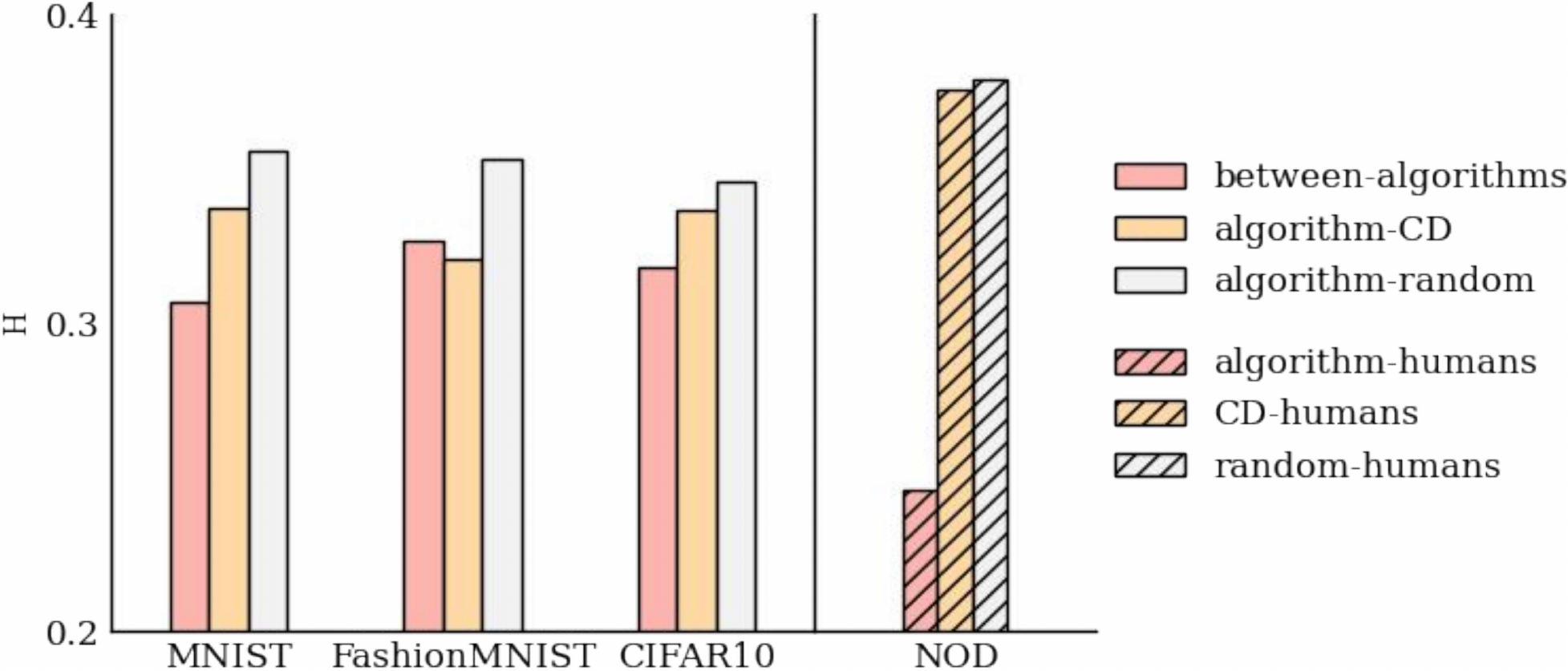}\vspace{-2mm}
  \caption{\textbf{There exists low discrepancy on optimal curricula determined by between-algorithms, algorithm-CD, algorithm-humans, and CD-humans.} 
  Left panel: curricula discrepancy $\mathcal{H}$ (\textbf{Sec~\ref{sec:eval}}) is reported between pairs of CL algorithms $\mathcal{A}$s (between-algorithm, blue), between $\mathcal{A}$s and our CD (algorithm-CD, green), between $\mathcal{A}$ and the random designer (algorithm-random, orange).
  Right panel: $\mathcal{H}$ is reported on NOD dataset between $\mathcal{A}$s and humans (algorithm-human, blue hashed), between CD and humans (CD-humans, green hashed), and between the random designer and humans (random-humans, orange hashed) (\textbf{Sec~\ref{sec:results.4}}). 
  }\vspace{-5mm} 
\label{fig:fig5}\vspace{-5mm}
\end{center}
\end{figure}


\subsection{Analysis on Curriculum Agreement}\label{sec:results.4}
We set out to study the extent of agreement among curricula empirically optimized for individual students. For example, do the most effective curricula for EWC share commonalities with the most effective curricula for LwF? To address this question, we report the discrepancy $\mathcal{H}$ between any sets of ranked curricula determined empirically by CL algorithms $\mathcal{A}$, by our CD, and by the random curriculum designer on three image datasets of varying complexity (\textbf{Fig~\ref{fig:fig5}}). A decrease in $\mathcal{H}$ indicates an increase in the agreement (\textbf{Sec~\ref{sec:eval}}).
As a lower bound (``between-algorithms''), we first calculated the averaged discrepancy $\mathcal{H}$ over all pairs of $\mathcal{A}$s chosen among Vanilla, EWC, and LwF (\textbf{Sec~\ref{sec:algos}}). We consistently observe a large $\mathcal{H}$ decrease in ``between-algorithms'' relative to ``algorithm-random'' (average discrepancy $\mathcal{H}$ between sets of empirically ranked curricula and set of randomly ranked curricula).
This implies that continual learning algorithms $\mathcal{A}$s agree with each other in empirically ranking the most effective curricula, more so than with random curricula. In other words, curricula that work well for one $\mathcal{A}$ tend to work well for another. 
We also examined the effect of $\mathcal{A}$'s hyperparameters on curriculum agreement (see \textbf{Sec~\ref{sec:training_regimes}} and \textbf{Fig~\ref{fig:fig_ablation}}), and found that the relative efficacy of curricula is consistent even with variations 
in the number of epochs, the learning rate and 
the network initialization.



We also assessed the discrepancy $\mathcal{H}$ between our CD's curriculum rankings and empirical curriculum rankings from $\mathcal{A}$. 
Across the three datasets (MNIST, FashionMNIST and CIFAR10, \textbf{Sec~\ref{sec:datasets}}), there is an average decrease of $0.02$ in $\mathcal{H}$ from algorithm-random to algorithm-CD.
It implies that our CD can predict optimal curricula well aligned with the curricula determined by $\mathcal{A}$s. However, $\mathcal{H}$ in between-algorithms is still higher than in algorithm-CD, indicating that the curricula ranked empirically by different $\mathcal{A}$s are more consistent with one another than with those ranked by our CD. 



The right panel in \textbf{Fig~\ref{fig:fig5}} shows the agreement in algorithm-humans, CD-humans, and random-humans
on the Novel Object Dataset (NOD, \textbf{Sec~\ref{sec:human_bench}}).  There is an $\mathcal{H}$ decrease of $0.13$ 
from random-humans to algorithm-humans. 
This indicates a notable degree of agreement between optimal curricula for humans and $\mathcal{A}$s. We further observe that there is a slight decrease in $\mathcal{H}$ 
from random-humans to CD-humans, indicating a minimal degree of alignment between humans and our CD. However, we notice that there still exists a huge gap in $\mathcal{H}$ from algorithm-humans to CD-humans. 

\section{Discussion}\label{sec:discussion}
\noindent 
Curriculum design is an important problem in both machine learning and human education. 
Key goals for both humans and machines include maximizing forward knowledge transfer across tasks while minimizing forgetting of previous tasks. In practice, there are numerous potential curriculum design considerations,
such as the ordering of training examples within and between classes and tasks, hierarchical learning across super-categories and sub-categories, learning characteristics of students, and feedback from students. 
Here, we introduce an initial proof-of-concept curriculum designer, which designs effective curricula for multiple CL algorithms by optimizing the ordering of a sequence of continuously learned tasks. 

While curriculum design proves effective for enhancing CL algorithms, its direct translation to human learning still encounters challenges.
To benchmark curriculum efficacy in humans, we introduced the Novel Object Dataset (NOD) and conducted human behavioral experiments. We observed a high discrepancy between optimal curricula ranked by our AI teacher and effective for human learning. There could be multiple reasons for this. First, the visual diets for humans and our AI teacher are different. Humans learn from temporally correlated video streams, which our AI teacher does not take 
into account. Second, there remains a gap between the background knowledge of humans and our AI teacher. Humans accumulate rich experiences through interactions with the real world involving multiple sensory modalities, but our AI teacher has been limited to knowledge from static naturalistic images in vision. Third, human individuals have large variability in learning due to individual cognitive capabilities and knowledge backgrounds. 
Our AI teacher lacks specialized curriculum designs for learning in individual humans.     



To resemble a human learning process, we took initial efforts and formulated our study of curriculum learning 
in the online class-incremental learning setting. 
Given computational resource constraints, we only exhaustively and empirically surveyed the 5-class and 10-class incremental settings on 3 CL algorithms across 3 datasets (\textbf{Sec~\ref{sec:datasets}}). 
Additional studies could explore a wider range of problem settings, such as task-incremental learning
and long-range CL with many classes. 
As a preliminary follow-up, we 
explored the effect of curriculum on the problem of visual question answering in function incremental settings (\textbf{Sec~\ref{sec:vqa}}, \textbf{Fig~\ref{fig:fig.vqa}}). We also investigated offline class-incremental learning, allowing the CL models to make multiple passes over the data within each task (\textbf{Sec~\ref{sec:multiple_epochs}}, \textbf{Fig~\ref{fig:fig.top10_bot10_paradigmI}}). Moreover, we extended our online learning tests to replay-based CL approaches (\textbf{Sec~\ref{sec:replay}}, \textbf{Fig~\ref{fig:fig.replay}}). Throughout all of these experiments, we observe curriculum effects that persist across variations in problem settings, datasets, and continual learning algorithms. 

AI for education and education for AI remain open challenges.
Our study establishes a methodology for the community to evaluate and benchmark curriculum design approaches for both humans and AI. 
 The insights obtained from our work open doors to many research opportunities, such as AI-assisted learning and education systems for both AI and human students. 


\section*{Acknowledgments}
This research is supported by the National Research Foundation, Singapore under its AI Singapore Programme (AISG Award No: AISG2-RP-2021-025), its NRFF award NRF-NRFF15-2023-0001, the National Science Foundation under grant number NSF CCF 1231216, the National Institutes of Health under grant number NIH R01EY026025, and the National Institute of General Medical Sciences under award number T32GM144273. 
We also acknowledge Mengmi Zhang's Startup Grant from Agency for Science, Technology, and Research (A*STAR), and Early Career Investigatorship from Center for Frontier AI Research (CFAR), A*STAR.
The authors declare that they have no competing interests. The funders had no role in study design, data collection and analysis, the decision to publish, or the preparation of the manuscript.






\onecolumn
\newpage
\clearpage
\resumetocwriting

\renewcommand{\thesection}{S\arabic{section}}
\renewcommand{\thefigure}{S\arabic{figure}}
\renewcommand{\thetable}{S\arabic{table}}
\setcounter{figure}{0}
\setcounter{section}{0}
\setcounter{table}{0}




\renewcommand{\contentsname}{List of Supplementary Sections}
\tableofcontents

\renewcommand{\listfigurename}{List of Supplementary Figures}
\listoffigures


\clearpage

\label{file:supp_cr_arxiv}

\section{\quad Experiments with Human Subjects}
\label{sec:hum_benchmark}


\subsection{Psychophysics Experiments} \label{sec:supp_psychophysics}
\noindent We took three precautions to control data quality and ensure that subjects paid attention to the experiments.
\begin{enumerate}[noitemsep,nolistsep]
    \item Subjects had to click on randomly presented triangles during the training rounds, and their reaction times were recorded for attention checks.
    \item  Subjects had to recognize simple geometric shapes, such as 3D cubes, in 
randomly dispersed dummy trials during the testing rounds.
\item In each testing round trial, the ``submit" button was disabled before the stimulus was shown for the full 200 millisecond presentation time to ensure that subjects were exposed to the stimulus. 
\end{enumerate}

\noindent For both MTurk and in-lab experiments, our results only incorporate data from subjects with 100\% accuracy in recognition of geometric shapes.


\subsection{Mechanical Turk experiments}

In our Amazon Mechanical Turk (MTurk) experiments, we collected responses from ``master workers'' with at least 1,000 approved human intelligence tasks (HITs) and a 95\% approval rate. We collected responses from 242 subjects in total. After filtering subjects for data quality (\textbf{Sec~\ref{sec:supp_psychophysics}}), we retained 169 subjects with 2-4 subjects for each tested curriculum. 

\begin{figure*}[htbp]
\begin{center}
\includegraphics[width=\textwidth]
{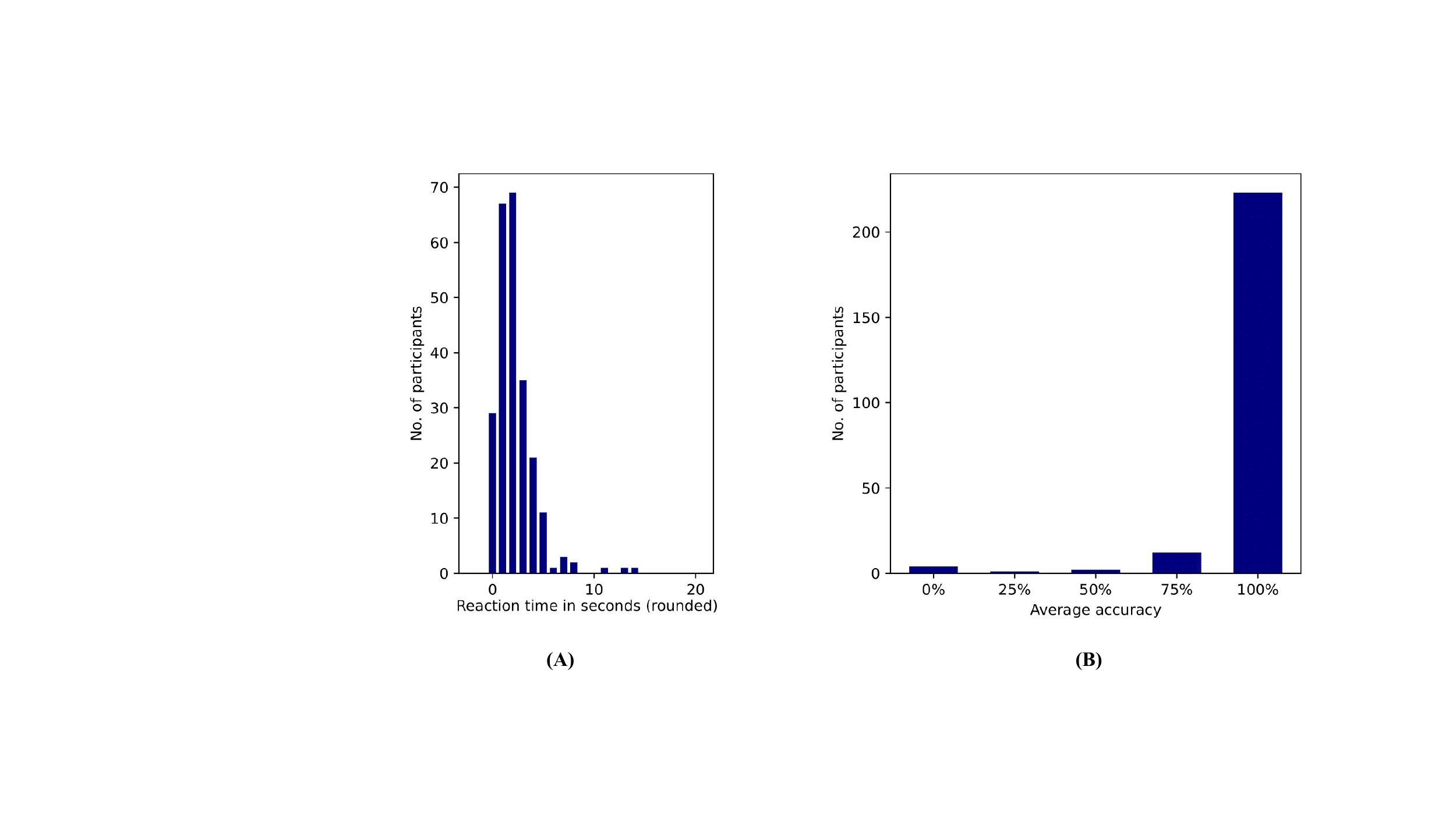}\vspace{-4mm}
\end{center}
  \caption[\textbf{Reaction time and attention check accuracy histograms for MTurk experiments}]{\textbf{Reaction time and attention check accuracy histograms for MTurk experiments.} (A) Reaction time distribution for all subjects in attention checks during training rounds. subjects were required to click on randomly presented triangles during the training rounds and their reaction time was recorded. 
  (B) Average accuracy of all subjects on attention checks during testing rounds. 
  We only used data from subjects who satisfied the criteria delineated in (\textbf{Sec~\ref{sec:hum_benchmark}}).
  }
\label{fig:exp_stats_mturk}
\end{figure*}

In \textbf{Fig~\ref{fig:exp_stats_mturk}A}, we show the distribution of reaction times from the attention checks for all MTurk subjects. We show the accuracy histogram of subjects on attention check trials in \textbf{Fig~\ref{fig:exp_stats_mturk}B}.

\begin{figure*}[htbp]
\begin{center}
\includegraphics[width=\textwidth]
{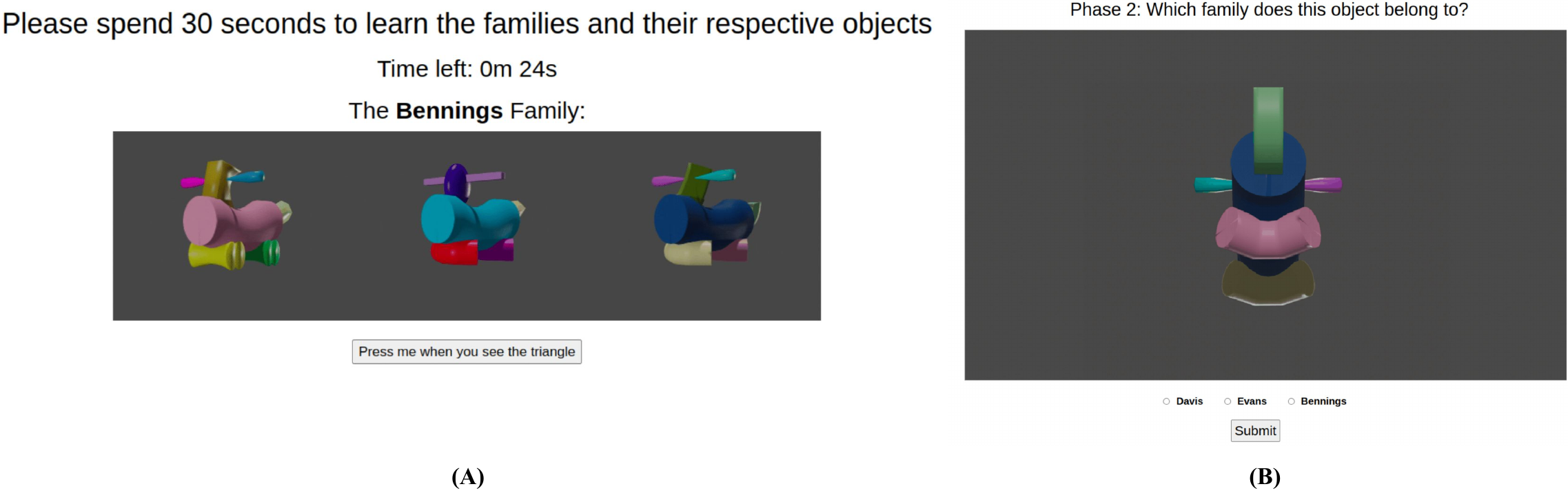}\vspace{-2mm}
\end{center}
  \caption[\textbf{MTurk interface schematics}]{\textbf{MTurk interface schematics.} Screenshots of the MTurk interface during the training rounds (A) and testing rounds (B).}
\vspace{2mm}
\label{fig:amt_interface}
\end{figure*}

In \textbf{Fig~\ref{fig:amt_interface}A} and \textbf{Fig~\ref{fig:amt_interface}B}, we show screenshots of the MTurk interface during the training and testing rounds of our experiment respectively. The exact same procedures and computer interfaces were used in the in-lab experiments.

We show the average accuracy of MTurk subjects over all tasks and an $\alpha$ vs. $\beta$ (\textbf{Sec~\ref{sec:eval}}) distribution for the Novel Object Dataset (NOD) in \textbf{Fig~\ref{fig:fig.line.novelnet}A} alongside results for in-lab subjects and Vanilla, EWC, and LwF continual learning (CL) algorithms. We also show the $\mathcal{F}$-scores of the top-5 vs. worst-5 performing curricula
in \textbf{Fig~\ref{fig:fig.topbot.mturk.inlab}A} alongside the best and worst curricula for in-lab subjects and the same CL algorithms. Overall, we observe a large effect of curriculum on learning performance in MTurk subjects. As shown in \textbf{Fig~\ref{fig:fig.topbot.mturk.inlab}}, between the top-5 and worst-5 curricula for the MTurk experiments, $\mathcal{F}$ ranges from $1.82\pm 0.12$ to $0.60\pm 0.09$. This difference in $\mathcal{F}$ here is significant.



\subsection{In-lab Experiments}
We augmented our study with in-lab experiments alongside the MTurk experiments to provide an additional layer of quality control. The exact same 
computer interfaces and experimental procedures were used for MTurk and in-lab experiments (\textbf{Fig~\ref{fig:exp_stats_mturk}}). 

It was infeasible to access a pool of subjects large enough to exhaustively test all possible curricula in-lab. The in-lab experiments were conducted only on 6 curricula, 3 of which were among the top-5 curricula as determined in the MTurk experiments, and the other 3 of which were among the worst-5 curricula from the MTurk experiments. As shown in \textbf{Fig~\ref{fig:fig.top_bot.nod.mturk_inlab}} (see legend for naming conventions), these 6 curricula are: (`fb3', `fc1', `fa1', `fb1', `fa2'), (`fb1', `fc1', `fb3', `fa2', `fa1'), (`fa1', `fb3', `fb1', `fc1', `fa2'), (`fa1', `fa2', `fc1', `fb1', `fb3'), (`fa1', `fb3', `fc1', `fa2', `fb1'), and (`fb1', `fb3', `fc1', `fa2', `fa1').
We recruited 60 subjects for in-lab experiments (10 for each curriculum), all of whom met the data quality criteria outlined in \textbf{Sec~\ref{sec:supp_psychophysics}}.

We evaluated the $\mathcal{F}$ score (\textbf{Sec~\ref{sec:eval}}) for each of the 6 in-lab curricula and compared $\mathcal{F}$ scores between in-lab and MTurk cohorts. As shown in \textbf{Fig~\ref{fig:fig.topbot.mturk.inlab}}, between the top-3 and worst-3 curricula for the in-lab experiments, $\mathcal{F}$ ranges from $1.65\pm 0.19$ to $1.30\pm 0.03$. This difference in $\mathcal{F}$ aligns with our observations from the MTurk results, though unlike in the MTurk results the difference here is not statistically significant.
Additionally, as can be seen from the curriculum visualizations in \textbf{Fig~\ref{fig:fig.top_bot.nod.mturk_inlab}}, the best curricula from the MTurk experiments are not identical to the best curricula from the in-lab experiments. 
However, 2 out of the 3 top curricula from the in-lab experiments were among the top-5 curricula for MTurk subjects, and the second-worst curriculum for the in-lab subjects was also the second-worst for MTurk subjects. 

\section{\quad Additional Information on Datasets for Paradigm-II}
\label{sec:datasets_supp} 
We conducted our experiments using three datasets: MNIST \cite{lecun1998mnist}, FashionMNIST \cite{xiao2017fashion}, and CIFAR10 \cite{krizhevsky2009learning}. 
Each dataset consists of 10 object classes. If classes are learned one at a time, each curriculum is a permutation of 10 classes, resulting in more than $3e^6$ (10!) possible curricula per dataset. Running all possible curricula is not practical due to computational resource constraints. To mitigate this issue, we introduce two paradigms. In paradigm-I, we chose a subset of 5 classes for each dataset (this paradigm produced the results described in the main paper, see \textbf{Sec~\ref{sec:datasets}}). In paradigm-II, we chose 5 tasks with 2 fixed classes each. In both paradigms, the order of the exemplars within each task is fixed and only the task sequence is permuted, resulting in a total of 5! $=$ 120 curricula.
The pair-wise groupings of the 10-classes from each dataset for paradigm-II was as follows:

\noindent \textbf{MNIST: } (`$0$,' `$1$'), (`$2$,' `$3$'), (`$4$,' `$5$'), (`$6$,' `$7$'), (`$8$,' `$9$').

\noindent \textbf{FashionMNIST: }(`shirt,' `sneaker'), (`top,' `trouser'), (`bag,' `boot'), (`coat,' `sandal'), (`pullover,' `dress').

\noindent \textbf{CIFAR10: }(`airplane,' `automobile'), (`frog,' `horse'), (`deer,' `dog'), (`ship,' `truck'), (`bird,' `cat')

\section{\quad Analysis Across Experimental Settings}
\label{sec:training_regimes}


We explored whether empirical performance discrepancies among curricula were consistent across experimental settings, specifically the number of epochs, parameter initialization procedures, and learning rates. 


For each experimental setting, we report the mean difference in curriculum discrepancy $\mathcal{H}$ (\textbf{Sec~\ref{sec:curriculum_discrepancy}}) among all pairs of CL algorithms $\mathcal{A}$s (between-algorithm) and between $\mathcal{A}$s and the random curriculum designer (algorithm-random) on FashionMNIST (\textbf{Fig~\ref{fig:fig_ablation}}). We vary only one experimental setting in each controlled experiment. 


First, we varied the number of training epochs over 1, 10, and 20 per incremental step for all $\mathcal{A}$s. Curriculum discrepancy $\mathcal{H}$ was $0.16$ lower on average in between-algorithms than in algorithm-random  over all three CL algorithms (\textbf{Fig~\ref{fig:fig_ablation}A}). This suggests that the relative efficacy of different curricula is similar regardless of whether algorithms train for one or multiple epochs.


Next, we vary the learning rates of all CL algorithms over $0.5e^{-3},\ 1e^{-3},$ and $2e^{-3}$. We observe $\mathcal{H}$ values that are lower by $0.02$ on average in between-algorithms comparisons than in algorithm-random comparisons (\textbf{Fig~\ref{fig:fig_ablation}B}).  
However, at the highest learning rate of $2e^{-3}$, the difference is much smaller than at lower learning rates. This suggests the hypothesis that, at high learning rates, curriculum effects may be either less impactful or less consistent in terms of which curricula are optimal.

Lastly, we tried several different network parameter initialization procedures: Gaussian, Uniform, and Xavier~\cite{glorot2010understanding}. We observed an average decrease of $0.03$ in the curriculum discrepancy from algorithm-random to between-algorithms (\textbf{Fig~\ref{fig:fig_ablation}C}). However, this decrease is much smaller for Xavier initialization than for the other two initialization procedures, suggesting that the extent to which optimal curricula agree across CL algorithms is dependent on the choice of parameter initialization procedure in at least some cases. 

\section{\quad Curriculum Affects Learning Performance Across Algorithms, Datasets, and Paradigms} 
\label{sec:curriculum_impact}


We analyze curriculum effects for three continual learning algorithms (\textbf{Sec~\ref{sec:algos}}) on three image datasets in both paradigm-I and paradigm-II (\textbf{Sec~\ref{sec:datasets}}).
For each analysis, we provide $\alpha$ versus $\beta$ plots (\textbf{Fig~\ref{fig:fig.line.mnist.5}}-\textbf{\ref{fig:fig.f.10}}), and the $\mathcal{F}$ distribution for the top-10 and bottom-10 curricula (\textbf{Fig~\ref{fig:fig.top10_bot10_paradigmI}}, \textbf{\ref{fig:fig.top10_bot10_paradigmII}}). 
Overall, the results suggest that curriculum significantly impacts performance in online class-incremental CL. Across all 18 scenarios (3 CL algorithms $\times$ 3 datasets $\times$ 2 paradigms), we observe statistically significant differences in performance between the 10 best and 10 worst curricula.  

\section{\quad Learning Effectiveness $\mathcal{F}$ as a Function of Time}
\label{sec:f_overtime}

We present the task-wise $\mathcal{F}$ score (\textbf{Sec~\ref{sec:eval}}) of the Vanilla CL algorithm, a ``random'' model, and an ``overfitting'' model (\textbf{Fig~\ref{fig:fig.f_overtime}}) across three datasets for paradigm-I (\textbf{Sec~\ref{sec:eval}}). In each task, the random model makes a random guess of the class label out of all the learned classes. The theoretical over-fitting model has perfect accuracy on the current task but has 100\% catastrophic forgetting and 0\% accuracy on previous tasks. We observe that the variance of $\mathcal{F}$ increases with increasing task number, implying a stronger curriculum effect with longer task sequences. We also observe that, even for the Vanilla algorithm, an effective curriculum leads to higher $\mathcal{F}$ than the overfitting and random models. Note that the overfitting model completely forgets task 1 when learning task 2; thus, $\mathcal{F}_{T=2}=2/(1-0 + 1/0.5)=0.67$ which is less than chance prediction. In case of chance, since each class would be assigned equal probability, we would have $\mathcal{F}_{T=2}=2/(1-0.5 + 1/0.5)=0.8$.


\section{\quad Alternative Curriculum Ranking Agreement Metric: Spearman's Rank Correlation Coefficient}
\label{sec:spearman}

As referenced in \textbf{Sec~\ref{sec:eval}}, we also calculate Spearman's rank correlation coefficients for curriculum ranking agreements on MNIST in paradigm-I (\textbf{Sec~\ref{sec:datasets}}), showing that it leads to the same conclusions as those reached using $\mathcal{H}$. We calculated Spearman's correlation coefficients of 0.26, 0.08, and 0.0002 for between-algorithms, algorithm-CD, and algorithm-random comparisons for MNIST in paradigm-I (averaging among pairs of CL algorithms $A$s). These findings are consistent with those in \textbf{Sec~\ref{sec:results.4}} based on $\mathcal{H}$: CL algorithms agree to a significant extent on empirical rankings of curricula, and our CD predicts these empirical rankings better than a random CD.


\section{\quad Our CD Predicts Optimal Curricula in Paradigm-II Based on Recall@K Measurements}
\label{sec:optimal_curricula}


Following the same figure interpretation as for paradigm-I in \textbf{Fig~\ref{fig:fig3}}, we report Recall@K results for paradigm-II in \textbf{Fig~\ref{fig:fig.recall@allk_10}}.
We found that our CD predicted optimal curricula more accurately than the random model on average across all three datasets, particularly at larger values of k. 
Moreover, we see no clear evidence that the performance of our CD is dependent on the difficulty of the classification tasks to be learned, since it performs well across three datasets with varying complexity.

\section{\quad Analysis of Curriculum Discrepancy in Paradigm-II}\label{sec:curricula_agreement}

\textbf{Fig~\ref{fig:fig.curricula.agreement}} illustrates the discrepancy $\mathcal{H}$ between curriculum rankings determined empirically by CL algorithms, heuristically by our curriculum designer (CD), and randomly by the random curriculum designer on MNIST, FashionMNIST, and CIFAR10 (\textbf{Sec~\ref{sec:datasets}}) in paradigm-II (10 classes arranged in 5 binary tasks, \textbf{Sec~\ref{sec:datasets}}). A decrease in $\mathcal{H}$ indicates an increase in the agreement between curriculum rankings (\textbf{Sec~\ref{sec:eval}}).

Like in Paradigm-I, we conclude that CL algorithms share a comparable set of top-ranked curricula across three datasets in Paradigm-II. We also assess curriculum agreement between our CD and CL algorithms. We observe an decrease of $0.01$ in the discrepancy from algorithm-random to algorithm-CD in CIFAR10. However, our CD fails for MNIST and FashionMNIST, yielding higher curriculum discrepancy with empirically ranked curricula than a random CD, despite identifying optimal curricula better than a random CD according to Recall@K (\textbf{Sec~\ref{sec:optimal_curricula}, Fig~\ref{fig:fig.recall@allk_10}}). This suggests that although our CD identified the highest-performing curricula relatively well for MNIST and FashionMNIST in this setting, it did not accurately predict the rankings of less effective curricula further down in the rankings.
In any case, there is still a great deal of room for improvement in predicting optimal curricula across datasets, algorithms, and training regimens.

\section{\quad CD Ablation Study in Paradigm-II}
\label{sec:cd.design}

We report the effects of ablating several CD design decisions in Paradigm-I in \textbf{Fig~\ref{fig:fig4}}, and repeat them in \textbf{Fig~\ref{fig:fig.recall@k}A} for convenience. \textbf{Fig~\ref{fig:fig.recall@k}B} shows CD ablation results for paradigm-II. We follow the same figure conventions as \textbf{Fig~\ref{fig:fig4}}. Unlike in the results from paradigm-I (see \textbf{Sec~\ref{sec:results.3}}), we did not observe clear benefits of our specific CD design choices in paradigm-II (e.g., as indicated by zero recall at k=5).

\section{\quad Curriculum Influences Performance of a Naive Replay Algorithm in Class-Incremental Online CL}
\label{sec:replay}

To extend our study of class-incremental online CL with Vanilla, EWC, and LwF, we investigate the effects of curricula on a naive replay CL algorithm. This algorithm used a replay buffer size equivalent to 10\% of the training set of each task (for example, if the training set comprised x images per task, the buffer size would $0.1x$) and adopted a random sampling strategy to select samples for the memory buffer. We did not experiment with the ordering of the replayed examples themselves. 



As observed in \textbf{Fig~\ref{fig:fig.replay}}, for MNIST, the average $\mathcal{F}$ scores ($\pm$ standard deviation) were $1.55\pm 0.06$ and $0.93\pm 0.04$ for the top-10 worst-10 curricula respectively. 
For FashionMNIST the average $\mathcal{F}$ scores were $1.26\pm 0.07$ and $0.79\pm 0.04$, and for CIFAR10 they were $1.14\pm 0.04$ and $0.63\pm 0.04$.
This suggests that curriculum plays a crucial role in the performance of replay-based continual learning algorithms. Across all three datasets, we observed that the top curricula outperform the worst curricula significantly.

We also assessed the curriculum discrepancy $\mathcal{H}$ (\textbf{Fig~\ref{fig:fig.replay}}) between pairs of curriculum rankings determined by CL algorithms (Vanilla, EWC, LwF and naive-replay; accounting for all pairs of CL algorithms) including the naive-replay CL algorithm, and a random curriculum designer. We observe a statistically significant decrease in $\mathcal{H}$ from $0.60\pm 0.001$ to $0.40\pm 0.07$ in $\mathcal{H}$ from between-algorithms to algorithm-random for MNIST.
For FashionMNIST, we observe a statistically significant $\mathcal{H}$ decrease from $0.60\pm 0.001$ to $0.40\pm 0.06$, 
and for CIFAR10 we observe a statistically significant $\mathcal{H}$ decrease from $0.62\pm 0.002$ to $0.38\pm 0.09$.
This implies that, compared to the agreement between the curricula ranked randomly and curricula ranked empirically by CL algorithms, the CL algorithms including naive-replay share comparable curriculum rankings.

\section{\quad Curriculum Influences Performance in Offline Class-Incremental Learning}
\label{sec:multiple_epochs}
We extend our investigation of curriculum effects in CL to offline class-incremental CL, where multiple passes over the data within each task are allowed. \textbf{Fig~\ref{fig:fig.top10_bot10_paradigmI}} highlights the effect of curricula on the Vanilla, EWC and LwF algorithms (\textbf{Sec~\ref{sec:algos}}) over three datasets (\textbf{Sec~\ref{sec:datasets}}) in this offline CL setting. Despite multi-epoch training on each task, the results are consistent with our findings as highlighted in \textbf{Sec~\ref{sec:results.1}} and \textbf{Sec~\ref{sec:curriculum_impact}}. 

\section{\quad Curriculum Strongly Affects Performance in Continual Visual Question Answering}
\label{sec:vqa}

To study the impact of curriculum in a multi-modal setting, we conducted additional experiments using Vanilla and EWC CL algorithms on the CLOVE VQA dataset (\cite{lei2022symbolic}). CLOVE is a benchmark dataset for CL in a VQA setting, and comprises question-answer (QA) pairs in five groups for function-incremental settings. The QA pairs are categorized based on the five functions of knowledge reasoning, object recognition, attribute recognition, relation reasoning, and logic reasoning.
Since computing the results across all possible curricula (5! = 120) was infeasible due to limited computational resources, we sampled 16 curricula at random. Despite only sampling a small subset of possible curricula, in \textbf{Fig~\ref{fig:fig.vqa}} we observe strong curriculum effects in the function incremental setting. $\mathcal{F}$ between the top-5 sampled curricula and worst-5 sampled curricula ranges from $0.64\pm 0.05$ to $0.36\pm 0.02$ using the Vanilla algorithm and ranges from $0.64\pm 0.04$ to $0.35\pm 0.02$ using the EWC algorithm 

We assessed the curriculum discrepancy $\mathcal{H}$ (\textbf{Fig~\ref{fig:fig.vqa}}) between pairs of curriculum rankings determined empirically by the Vanilla and EWC algorithms, and by a random curriculum designer. We observe a significant decrease in $\mathcal{H}$ from $0.76\pm 0.0004$ to $0.24\pm 0.0001$  
from algorithm-random to between-algorithms, indicating that the CL algorithms share a comparable set of top-ranked curricula when compared to the agreement between randomly and empirically ranked curricula.

It is intriguing that, in general, the curriculum effects we observe in VQA are dramatically larger than those we observe in image classification. This experiment further supports the conclusion that curriculum plays an important role in continual learning, perhaps especially in complex continual learning settings such as continual VQA.


\section{\quad Statistical Analysis}
\label{sec:stat_anal}

We employed two-sample t-tests to compute statistical significance in the following cases: (1) comparing the top-k $\mathcal{F}$ scores to the bottom-k $\mathcal{F}$ scores to establish the presence of curriculum effects (see \textbf{Sec~\ref{sec:hum_benchmark}}, \textbf{\ref{sec:curriculum_impact}}, \textbf{\ref{sec:vqa}}, \textbf{\ref{sec:replay}} and \textbf{Fig~\ref{fig:fig.topbot.mturk.inlab}}, \textbf{\ref{fig:fig.top10_bot10_paradigmI}}, \textbf{\ref{fig:fig.top10_bot10_paradigmII}}, \textbf{\ref{fig:fig.replay}}, \textbf{\ref{fig:fig.vqa}}), and (2) comparing two sets of $\mathcal{H}$ (\textbf{Fig~\ref{fig:fig.replay}}, \textbf{\ref{fig:fig.vqa}}) to discern if the curriculum agreement between two distributions varies significantly or not. 
We use the asterisk symbol * in all relevant figures to denote significant p-values ($p<0.05$) in 2-sample t-tests, and use ``n.s." to denote higher non-significant p-values. Errorbars are also presented to indicate standard deviation across all test trials. 




\pagebreak





\begin{figure*}[htbp]
\begin{center}
\includegraphics[width=\textwidth]
{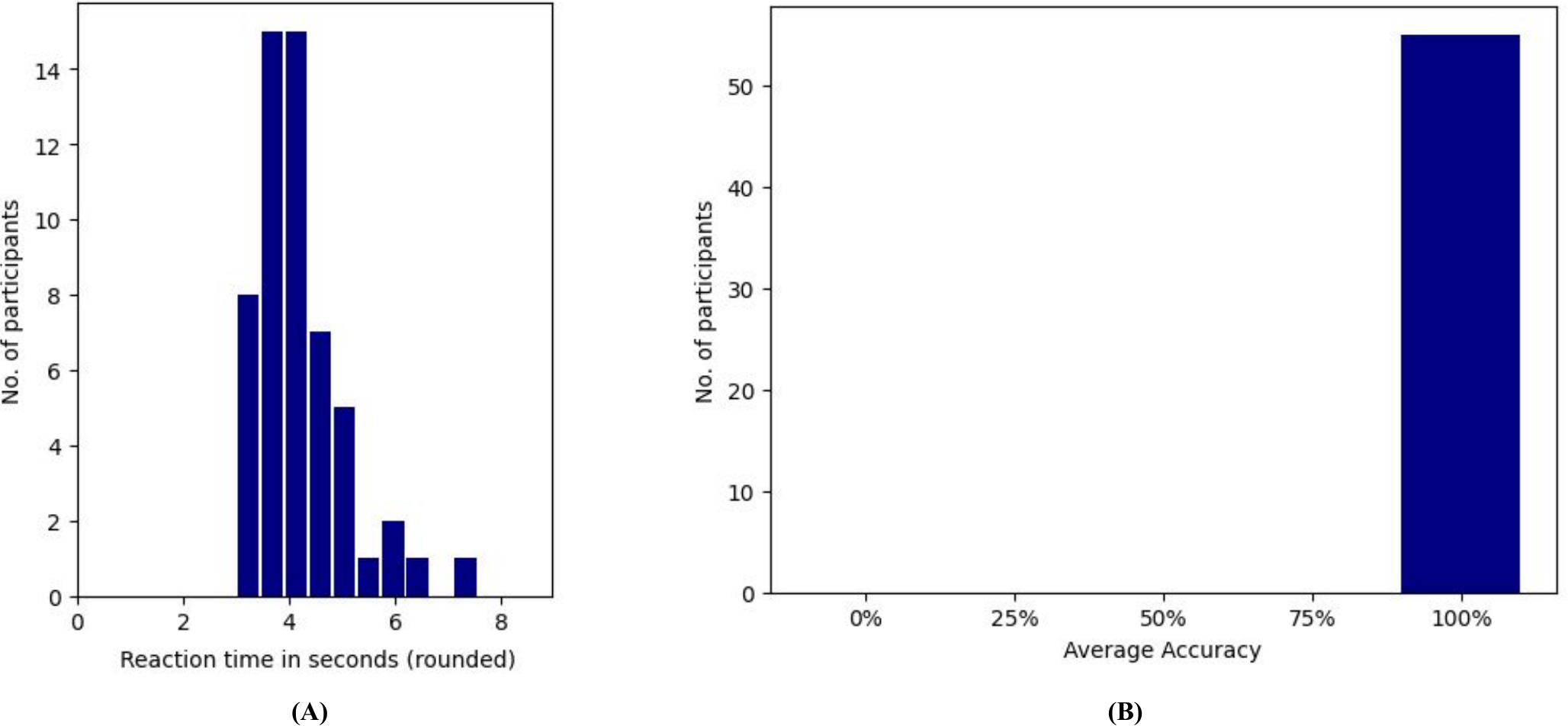}\vspace{-4mm}
\end{center}
  \caption[\textbf{Reaction time and attention check accuracy for in-lab experiments}]{\textbf{Reaction time and attention check accuracy for in-lab experiments.} (A) Reaction time distribution for all subjects in attention checks during training rounds. Subjects were required to click on randomly presented triangles during the training rounds and their reaction time was recorded. On the x-axis, we show the reaction time in seconds (rounded). (B) Average accuracy of all subjects in attention checks during testing rounds. 
  All in-lab subjects were included in our analysis, since all subjects' data satisfied the inclusion criteria in \textbf{Sec~\ref{sec:hum_benchmark}}.}
\label{fig:exp_stats_inlab}
\end{figure*}

\begin{figure*}[htbp]
\begin{center}
\includegraphics[width=0.5\textwidth]{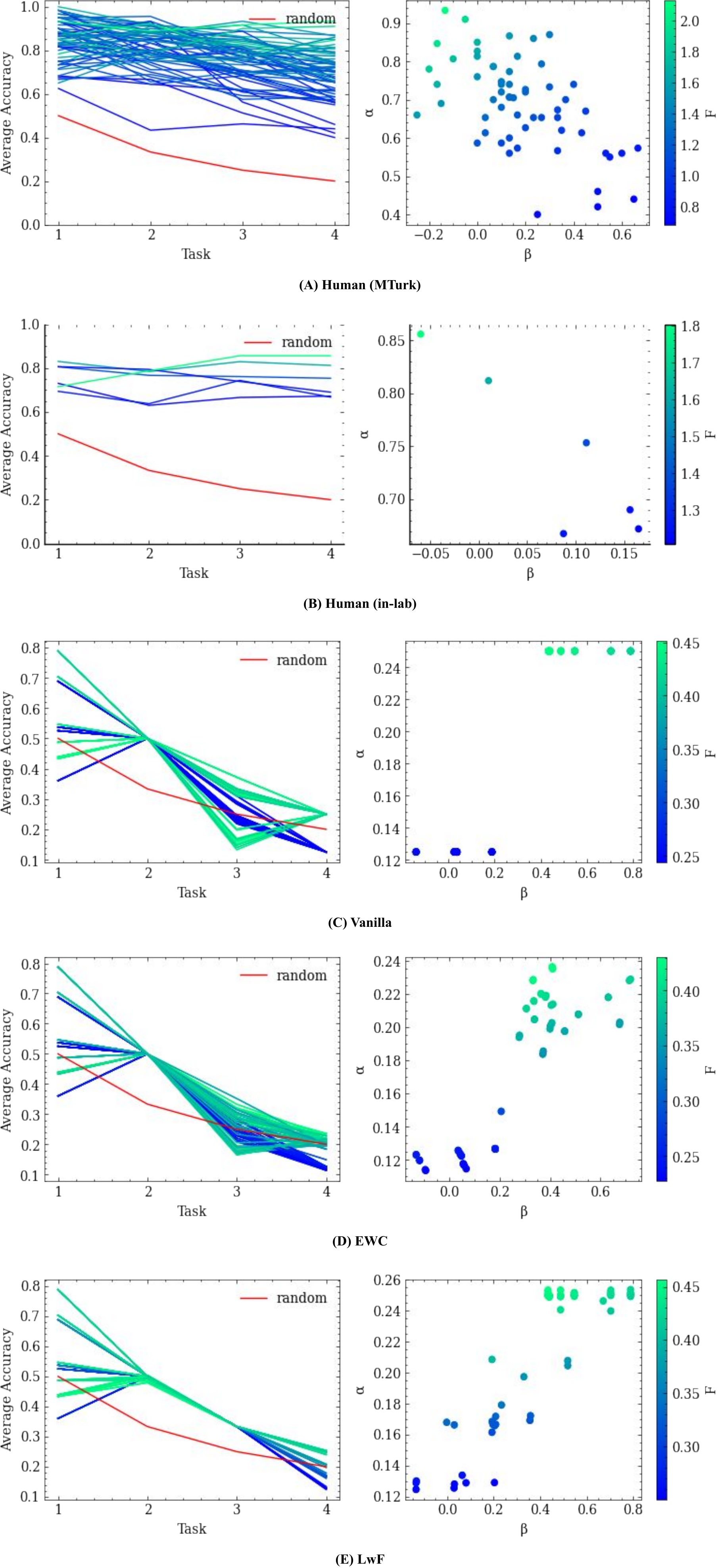}\vspace{-6mm}
\end{center}
  \caption[\textbf{Curriculum effects on performance on NOD for humans and for Vanilla, EWC and LwF CL algorithms}]{\textbf{Curriculum effects on performance on NOD for humans and for Vanilla, EWC and LwF CL algorithms (Sec~\ref{sec:algos}, Sec~\ref{sec:human_bench}).}  
  We report the average accuracy across all tasks in the left-hand panel for each condition. We also plot $\alpha$ vs $\beta$ (\textbf{Sec~\ref{sec:eval}}) in the right-hand panel for each condition. The effectiveness measure $\mathcal{F}$ (\textbf{Sec~\ref{sec:eval}}) incorporates both $\alpha$ and $\beta$.}
\label{fig:fig.line.novelnet}
\end{figure*}

\begin{figure*}[htbp]
\begin{center}
\includegraphics[width=\textwidth, scale=0.8]{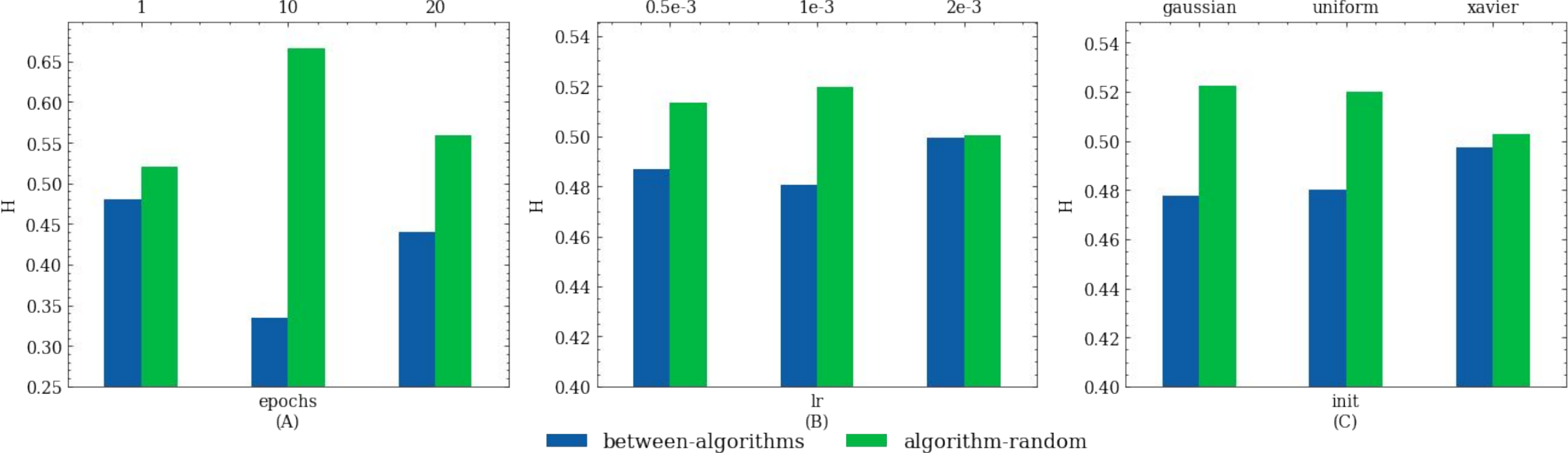}\vspace{-6mm}
\end{center}
  \caption[\textbf{Curriculum agreement (low curriculum discrepancy $\mathcal{H}$) among CL algorithms persists across different experimental settings on FashionMNIST}]{\textbf{Curriculum agreement (low curriculum discrepancy $\mathcal{H}$) among CL algorithms persists across different experimental settings on FashionMNIST.} The discrepancy between two sets of ranked curricula is measured as $\mathcal{H}$, with smaller values indicating lower discrepancy and higher curriculum agreement (\textbf{Sec~\ref{sec:curriculum_discrepancy}}). Within each pair of bars, the discrepancy between pairs of ranked curriculum sets for CL algorithms $\mathcal{A}$s (between-algorithms) is presented on the left (blue), and that between $\mathcal{A}$ and the randomly ranked curricula (algorithm-random) is on the right (green). We vary the number of epochs (A), the learning rates (lr) (B), and the network parameter initialization procedure (C). For visualization purposes, within each pair of bars, 
    we normalize the $\mathcal{H}$ value over between-algorithm and algorithm-random so that the sum of these two discrepancy values (green $+$ blue) always equals 1. Normalization does not alter the main conclusion that curriculum discrepancy is always lower in the between-algorithms condition, meaning the same curricula work well (and the same curricula work poorly) across a range of experimental conditions.
    }\vspace{-2mm}
\label{fig:fig_ablation}
\end{figure*}

\begin{figure*}[htbp]
\begin{center}
\includegraphics[width=\textwidth, scale=0.6]{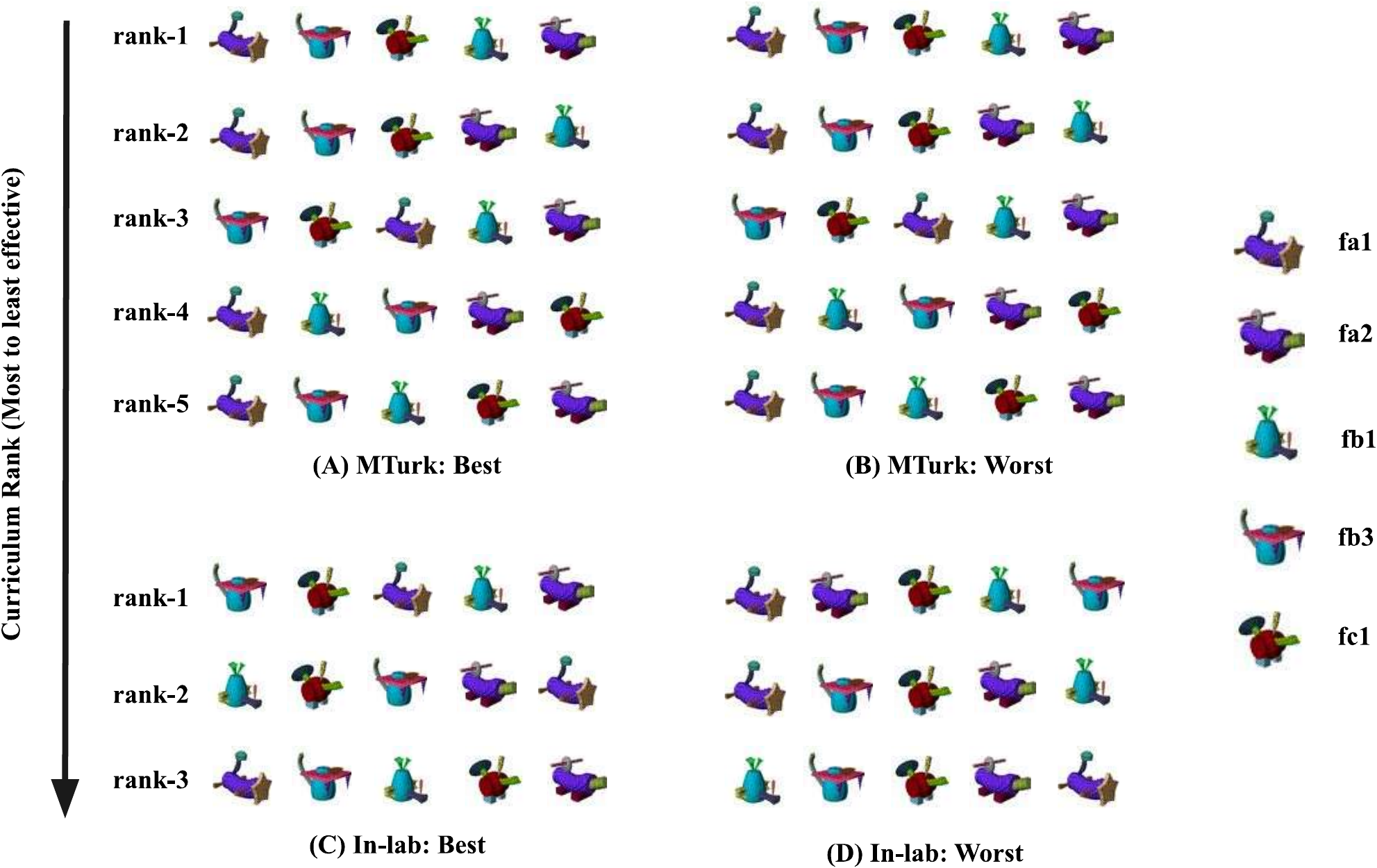}\vspace{-6mm}
\end{center}
  \caption[\textbf{Experimentally determined best and worst curricula on NOD for MTurk and in-lab human subjects}]{ \textbf{Experimentally determined best and worst curricula on NOD (Sec~\ref{sec:human_bench}) for MTurk (A-B, Sec~\ref{sec:human_bench}) and in-lab (C-D, Sec~\ref{sec:hum_benchmark}) human subjects.} Each row in the figure is one curriculum. The curricula are arranged from best to worst with the best curricula at the top.}\vspace{-2mm}
\label{fig:fig.top_bot.nod.mturk_inlab}
\end{figure*}

\begin{figure*}[htbp]
\begin{center}
\includegraphics[width=\textwidth, scale=0.8]{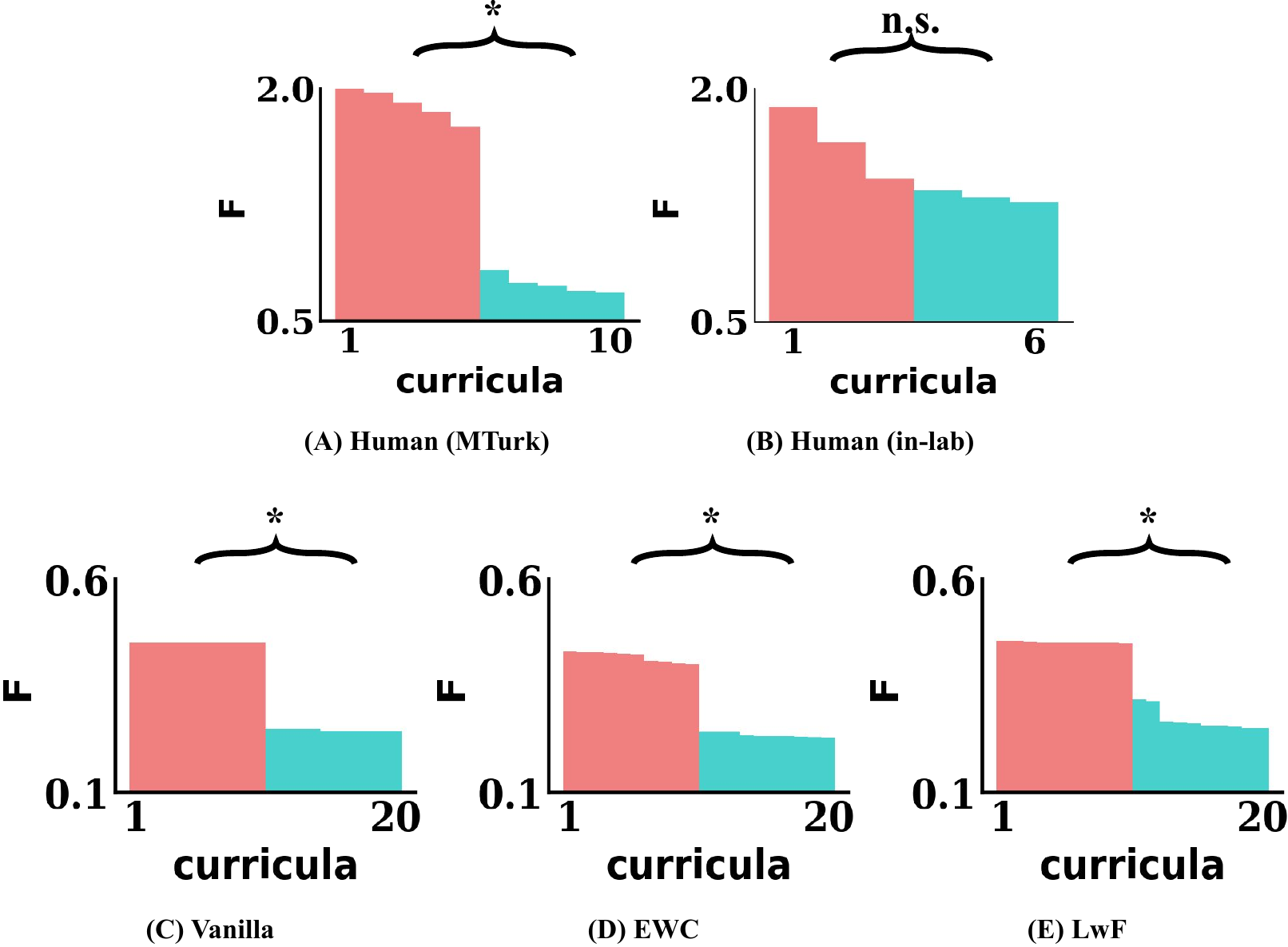}\vspace{-6mm}
\end{center}
   \caption[\textbf{Best and worst k curricula on NOD for MTurk and in-lab human subjects, and for Vanilla, EWC, and LwF CL algorithms}]{ 
   \textbf{Best and worst k curricula on NOD (Sec~\ref{sec:datasets}) for (A) MTurk human subjects (top 5 vs bottom 5), (B) in-lab human subjects (top 3 vs bottom 3), (C) Vanilla (top 10 vs bottom 10), (D) EWC (top 10 vs bottom 10), and (E) LwF (top 10 vs bottom 10)}. The plot shows the $\mathcal{F}$-scores for the best curricula (red) and the worst curricula (blue) as well as the statistical significance (* = statistically significant) determined via two-sample t-tests on the $\mathcal{F}$-scores of the best and worst curricula. 
  }\vspace{-2mm}
\label{fig:fig.topbot.mturk.inlab}
\end{figure*}

\begin{figure*}[htbp]
\begin{center}
\includegraphics[width=1\textwidth]{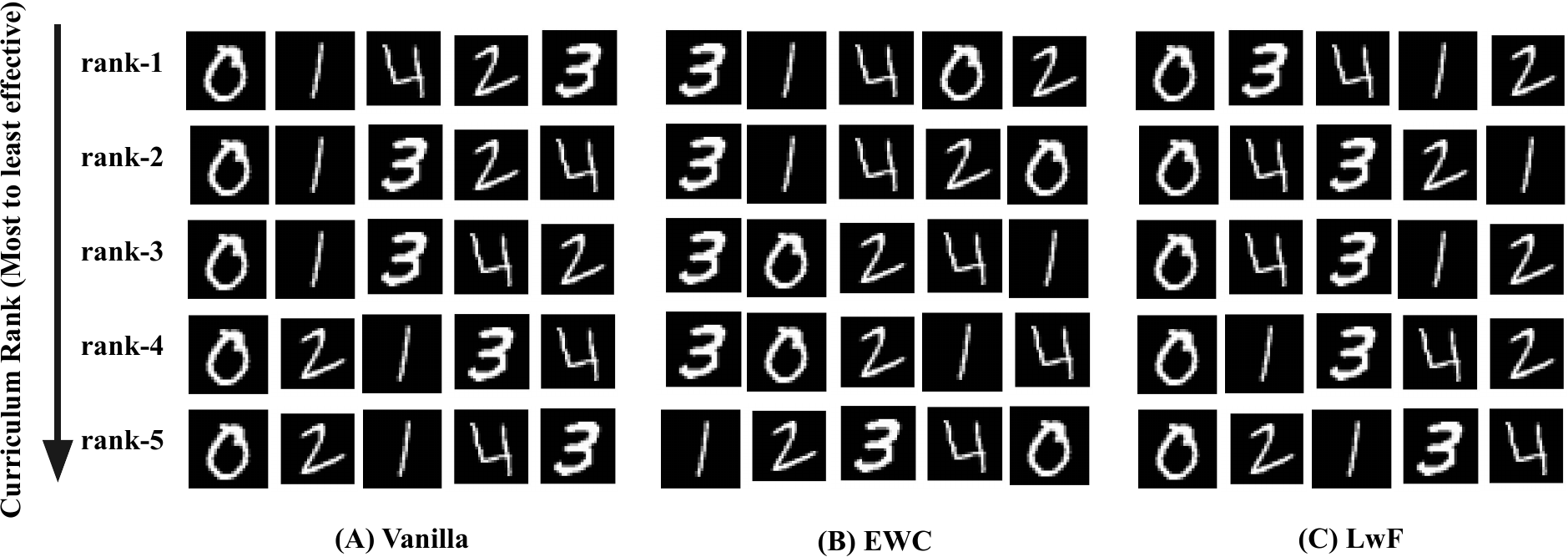}\vspace{-6mm}
\end{center}
  \caption[\textbf{Empirically determined top-5 curricula on MNIST for Vanilla, EWC and LwF CL algorithms in paradigm-I}]{\textbf{Empirically determined top-5 curricula on MNIST for Vanilla, EWC and LwF CL algorithms (Sec~\ref{sec:algos}) in paradigm-I (5 classes, Sec~\ref{sec:datasets}).} Each row in the figure is one curriculum. Curricula are in descending order of effectiveness, with the best curriculum at the top. 
  }\vspace{-2mm}
\label{fig:fig.top5.mnist}
\end{figure*}

\begin{figure*}[htbp]
\begin{center}
\includegraphics[width=1\textwidth]{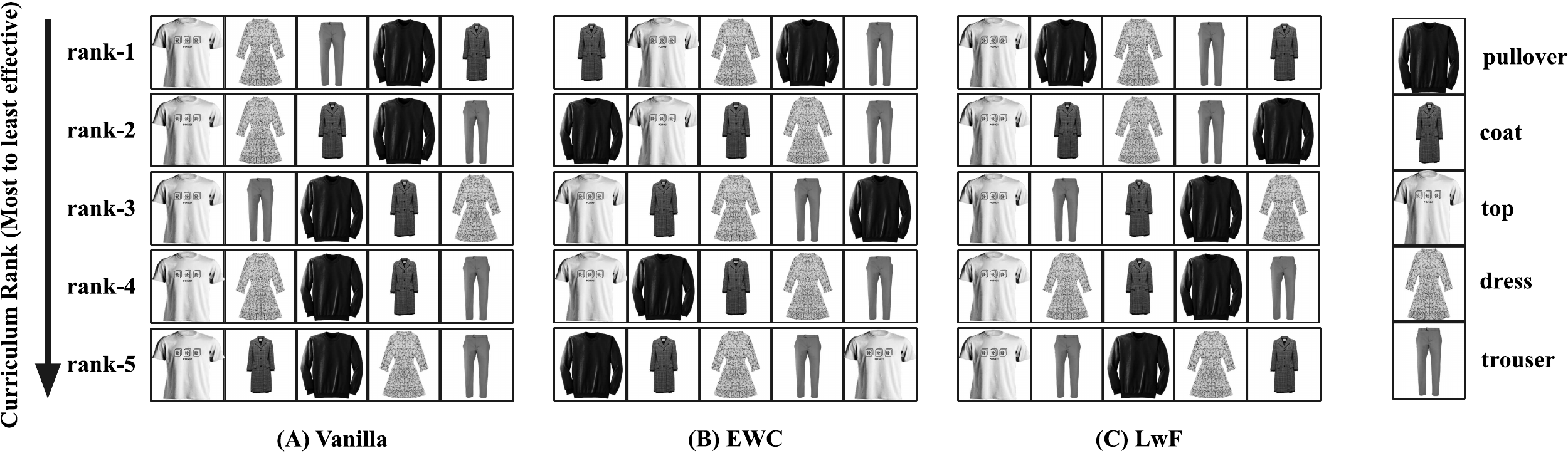}\vspace{-6mm}
\end{center}
  \caption[\textbf{Empirically determined top-5 curricula on FashionMNIST for Vanilla, EWC, and LwF CL algorithms in paradigm-I}]{ \textbf{Empirically determined top-5 curricula on FashionMNIST for Vanilla, EWC, and LwF CL algorithms (Sec~\ref{sec:algos}) in paradigm-I (5 classes, Sec~\ref{sec:datasets}).} Each row in the figure is one curriculum. Curricula are in descending order of effectiveness, with the best curriculum at the top.
  }\vspace{-2mm}
\label{fig:fig.top5.fmnist}
\end{figure*}

\begin{figure*}[htbp]
\begin{center}
\includegraphics[width=1\textwidth]{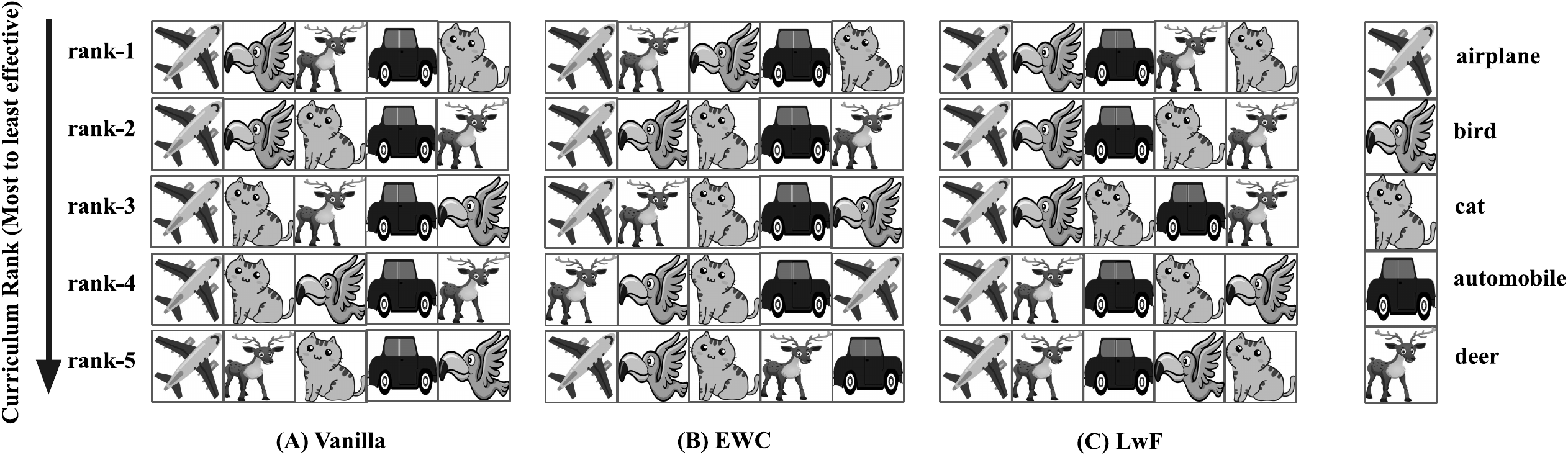}\vspace{-6mm}
\end{center}
  \caption[\textbf{Empirically determined top-5 curricula on CIFAR10 for Vanilla, EWC and LwF CL algorithms in paradigm-I}]{\textbf{Empirically determined top-5 curricula on CIFAR10 for Vanilla, EWC and LwF CL algorithms (Sec~\ref{sec:algos}) in paradigm-I (5 classes, Sec~\ref{sec:datasets}).} Each row in the figure is one curriculum. Curricula are in descending order of effectiveness, with the best curriculum at the top. For ease of interpretation, cartoon images are used to represent each class instead of actual CIFAR10 images.}\vspace{-2mm}
\label{fig:fig.top5.cifar}
\end{figure*}

\begin{figure*}[htbp]
\begin{center}
\includegraphics[width=1\textwidth]{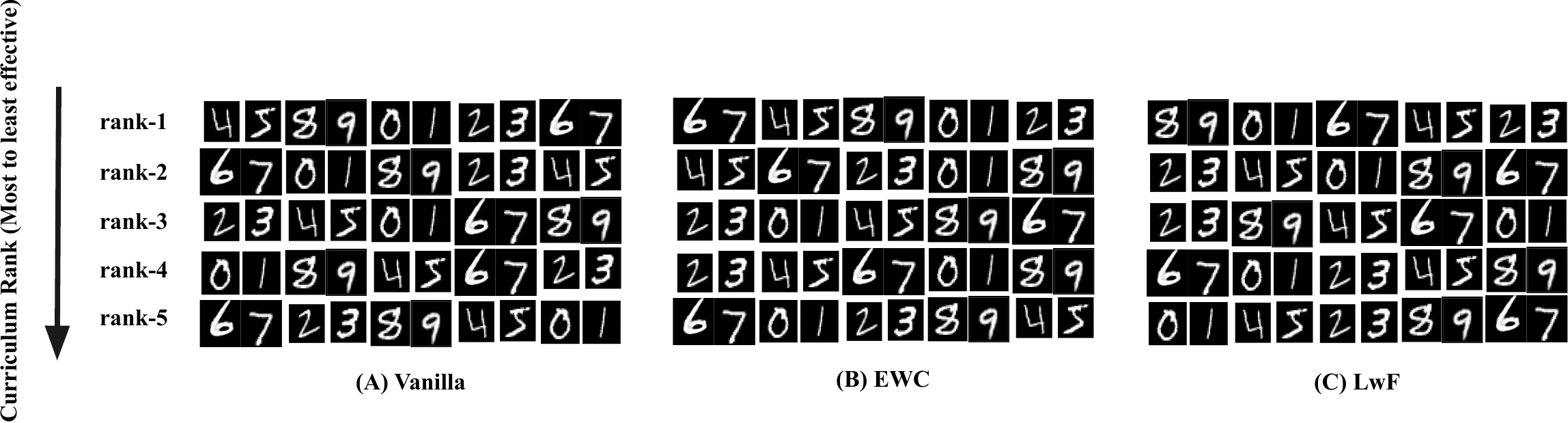}\vspace{-6mm}
\end{center}
  \caption[\textbf{Empirically determined top-5 curricula on MNIST for Vanilla, EWC, and LwF CL algorithms in paradigm-II}]{ \textbf{Empirically determined top-5 curricula on MNIST for Vanilla, EWC, and LwF CL algorithms (Sec~\ref{sec:algos}) in paradigm-II (10 classes arranged in 5 binary tasks, Sec~\ref{sec:datasets}).} Each row in the figure is one curriculum. Curricula are in descending order of effectiveness, with the best curriculum at the top.}\vspace{-2mm}
\label{fig:fig.top5.mnist.10}
\end{figure*}

\begin{figure*}[htbp]
\begin{center}
\includegraphics[width=1\textwidth]{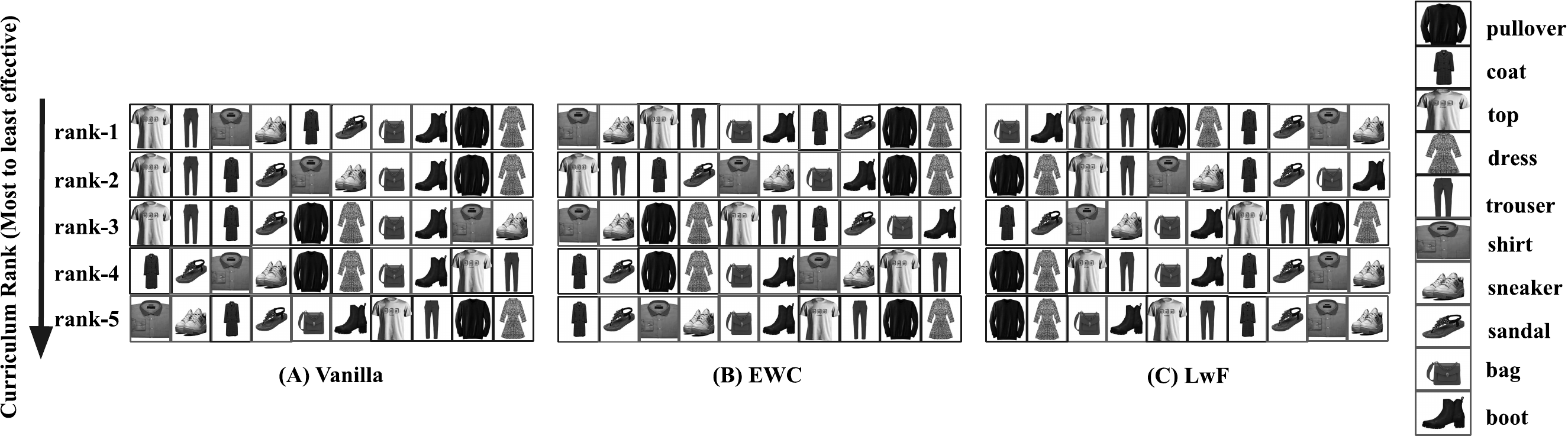}\vspace{-6mm}
\end{center}
  \caption[\textbf{Empirically determined top-5 curricula on FashionMNIST for Vanilla, EWC, and LwF CL algorithms in paradigm-II}]{ \textbf{Empirically determined top-5 curricula on FashionMNIST for Vanilla, EWC, and LwF CL algorithms (Sec~\ref{sec:algos}) in paradigm-II (10 classes arranged in 5 binary tasks, Sec~\ref{sec:datasets}).} Each row in the figure is one curriculum. Curricula are in descending order of effectiveness, with the best curriculum at the top.
  }\vspace{-2mm}
\label{fig:fig.top5.fmnist.10}
\end{figure*}

\begin{figure*}[htbp]
\begin{center}
\includegraphics[width=1\textwidth]{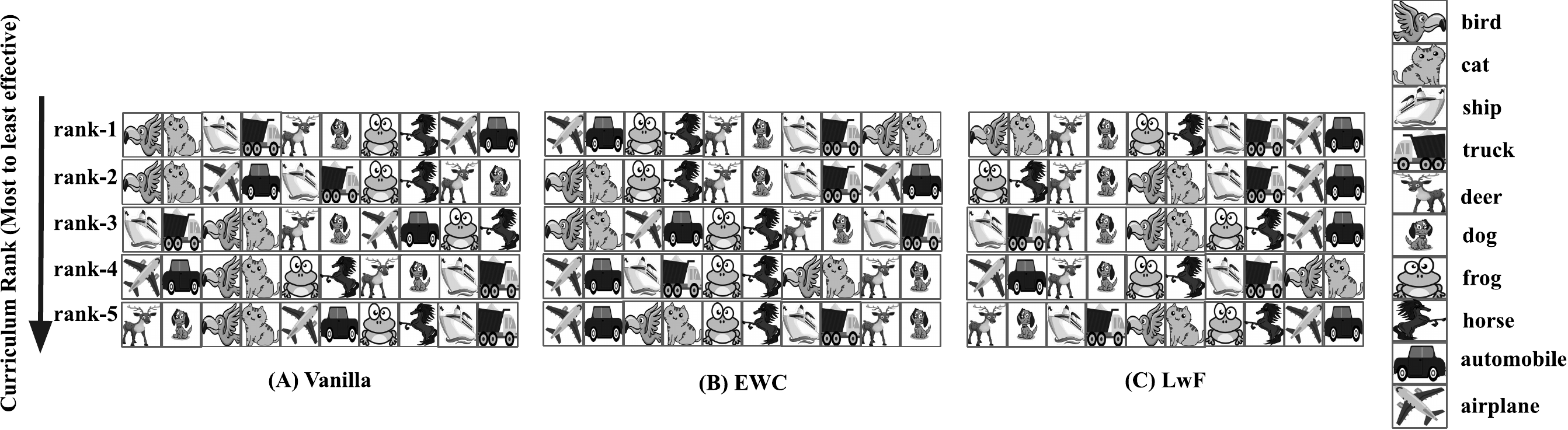}\vspace{-6mm}
\end{center}
  \caption[\textbf{Empirically determined top-5 curricula on CIFAR10 for Vanilla, EWC, and LwF CL algorithms in paradigm-II}]{ \textbf{Empirically determined top-5 curricula on CIFAR10 for Vanilla, EWC, and LwF CL algorithms (Sec~\ref{sec:algos}) in paradigm-II (10 classes arranged in 5 binary tasks, Sec~\ref{sec:datasets}).} Each row in the figure is one curriculum. Curricula are in descending order of effectiveness, with the best curriculum at the top. For ease of interpretation, cartoon images are used to represent each class instead of actual CIFAR10 images.}\vspace{-2mm}
\label{fig:fig.top5.cifar.10}
\end{figure*}

\clearpage

\begin{figure*}[htbp]
\begin{center}
\includegraphics[width=0.8\textwidth]{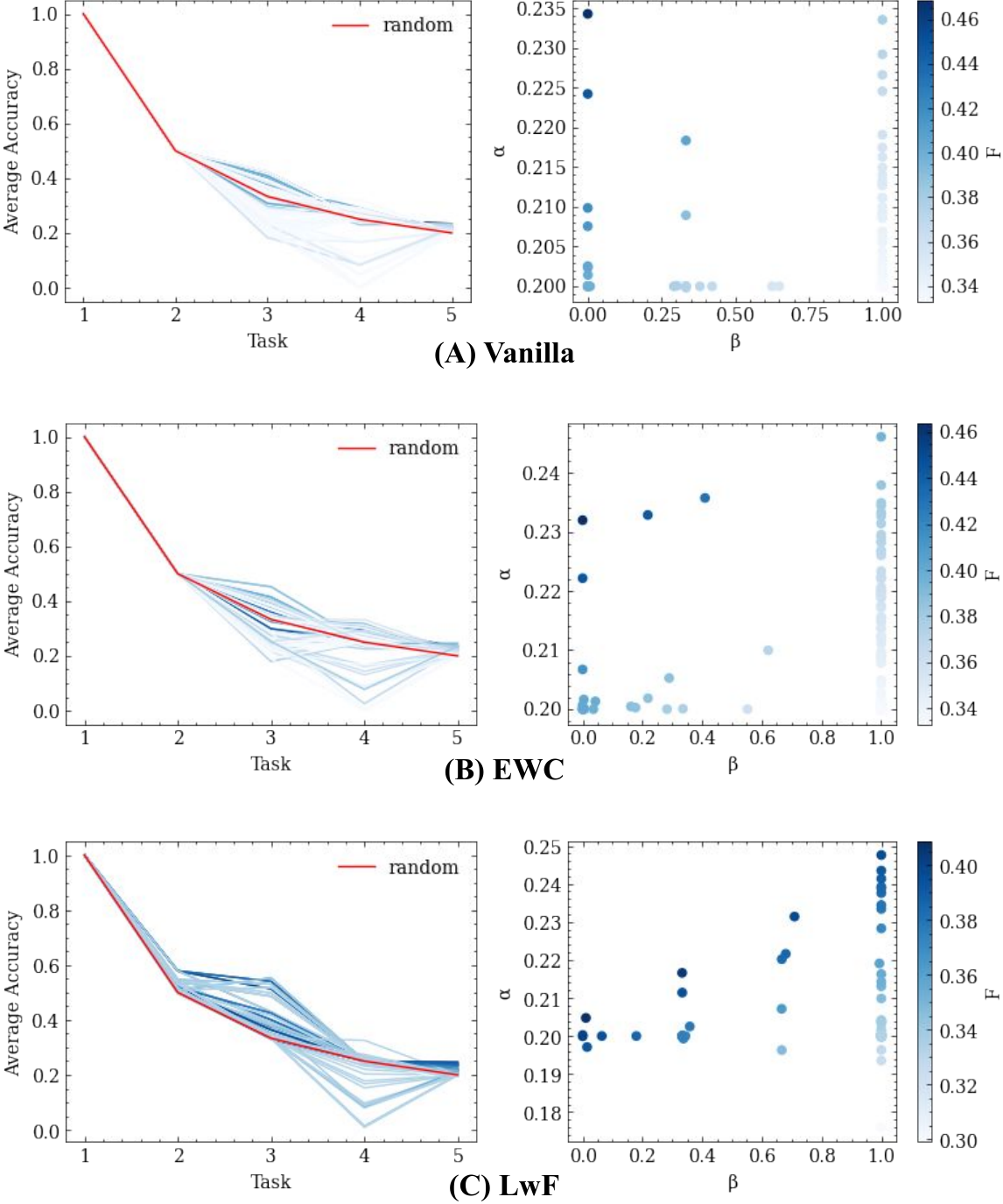}\vspace{-6mm}
\end{center}
  \caption[\textbf{Curriculum affects performance on MNIST for the Vanilla, EWC and LwF CL algorithms in paradigm-I}]{ \textbf{Curriculum affects performance on MNIST for the Vanilla, EWC and LwF CL algorithms (Sec~\ref{sec:algos}) in paradigm-I (5 classes, Sec~\ref{sec:datasets}).} This figure follows the same design conventions as \textbf{Fig~\ref{fig:fig.line.novelnet}}. 
  }
\label{fig:fig.line.mnist.5}
\end{figure*}

\begin{figure*}[htbp]
\begin{center}
\includegraphics[width=0.8\textwidth]{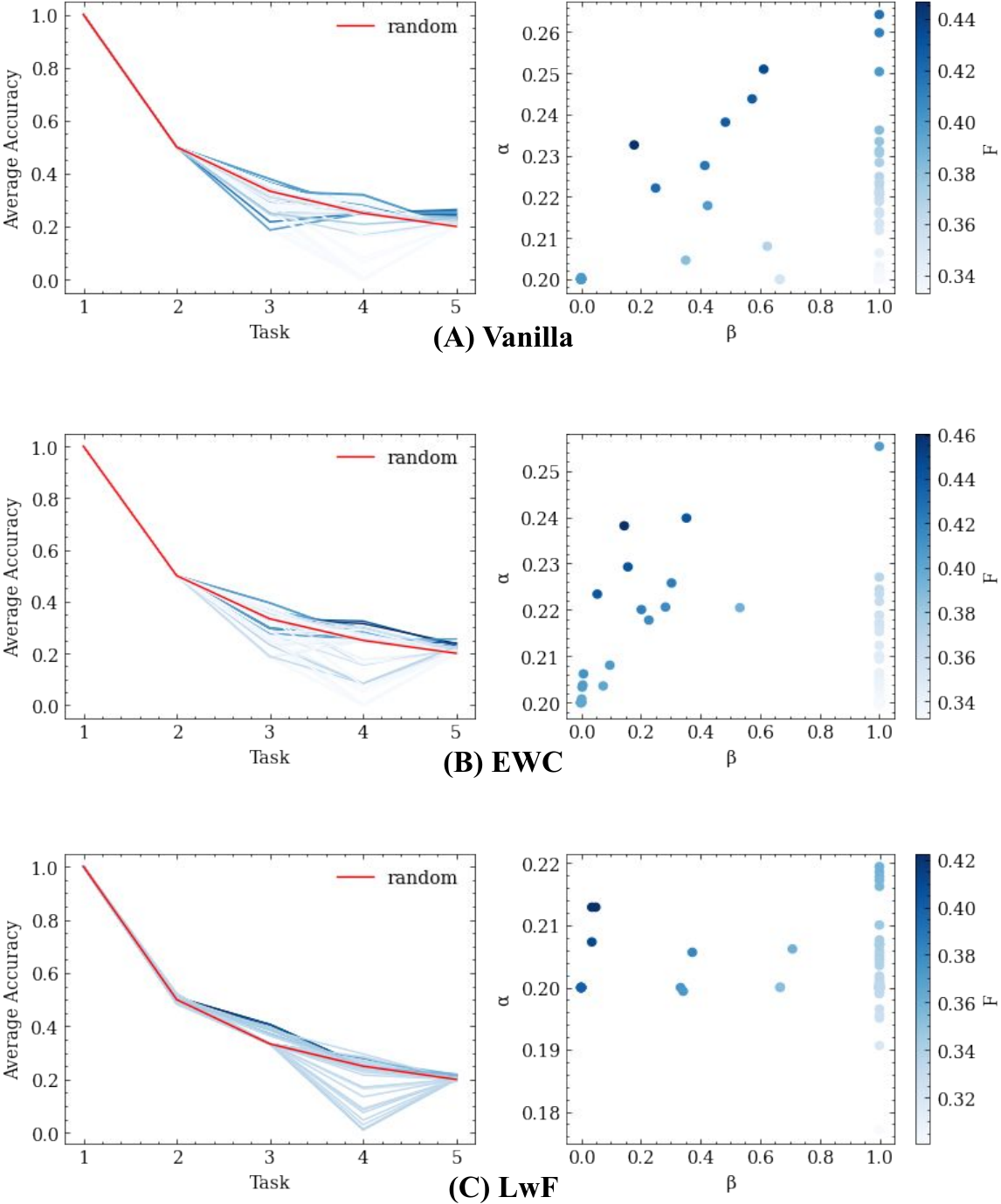}\vspace{-6mm}
\end{center}
  \caption[\textbf{Curriculum affects performance on FashionMNIST of the Vanilla, EWC and LwF CL algorithms in paradigm-I}]{ \textbf{Curriculum affects performance on FashionMNIST of the Vanilla, EWC and LwF CL algorithms (Sec~\ref{sec:algos}) in paradigm-I (5 classes, Sec~\ref{sec:datasets}).} This figure follows the same design conventions as \textbf{Fig~\ref{fig:fig.line.novelnet}}.
  }
\label{fig:fig.line.fmnist.5}
\end{figure*}

\begin{figure*}[htbp]
\begin{center}
\includegraphics[width=0.8\textwidth]{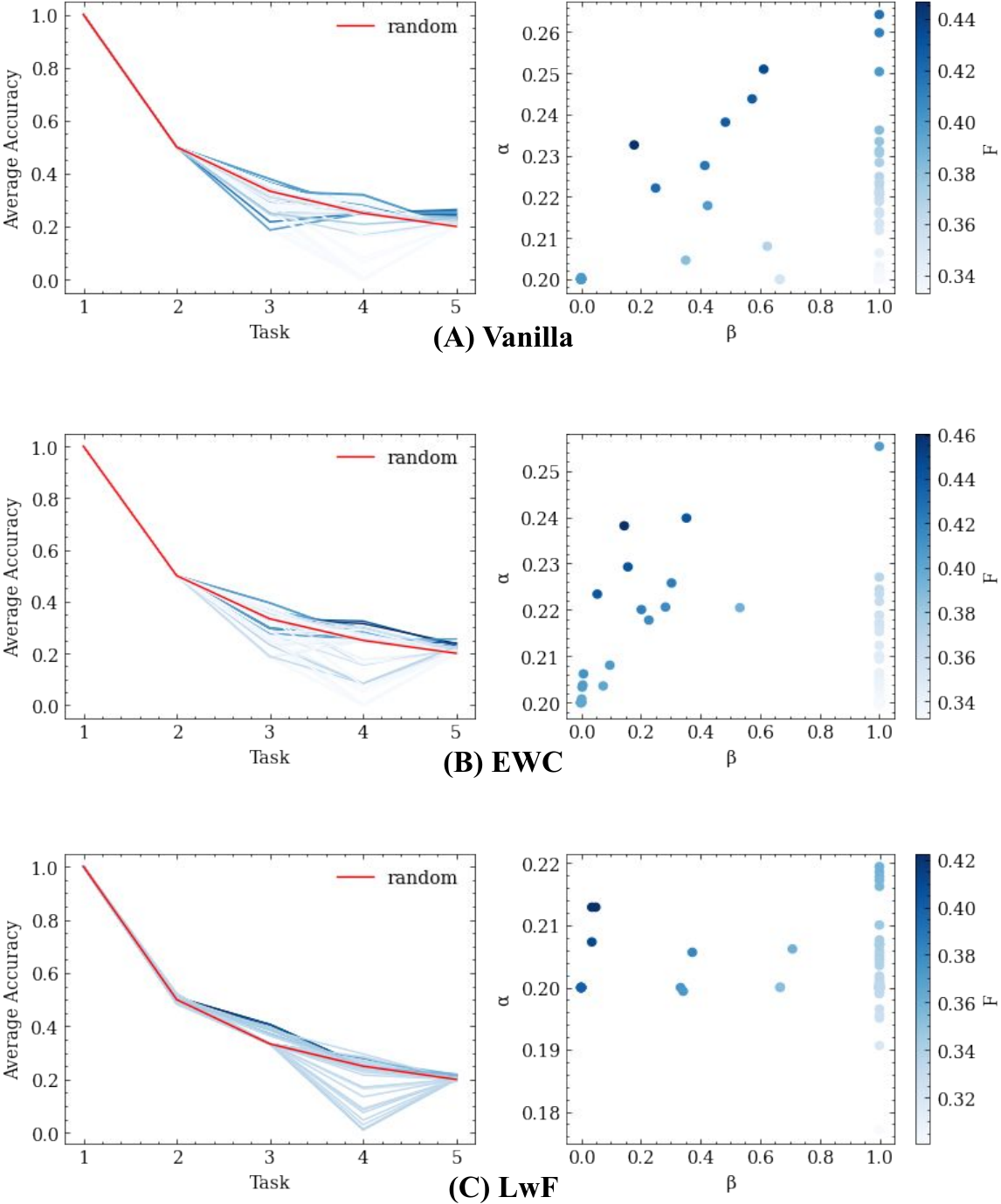}\vspace{-6mm}
\end{center}
  \caption[\textbf{Curriculum affects performance on CIFAR10 of the Vanilla, EWC and LwF CL algorithms in paradigm-I}]{ \textbf{Curriculum affects performance on CIFAR10 of the Vanilla, EWC and LwF CL algorithms (Sec~\ref{sec:algos}) in paradigm-I (5 classes, Sec~\ref{sec:datasets}).} This figure follows the same design conventions as \textbf{Fig~\ref{fig:fig.line.novelnet}}.
  }
\label{fig:fig.line.cifar.5}
\end{figure*}

\begin{figure*}[htbp]
\begin{center}
\includegraphics[width=\textwidth]{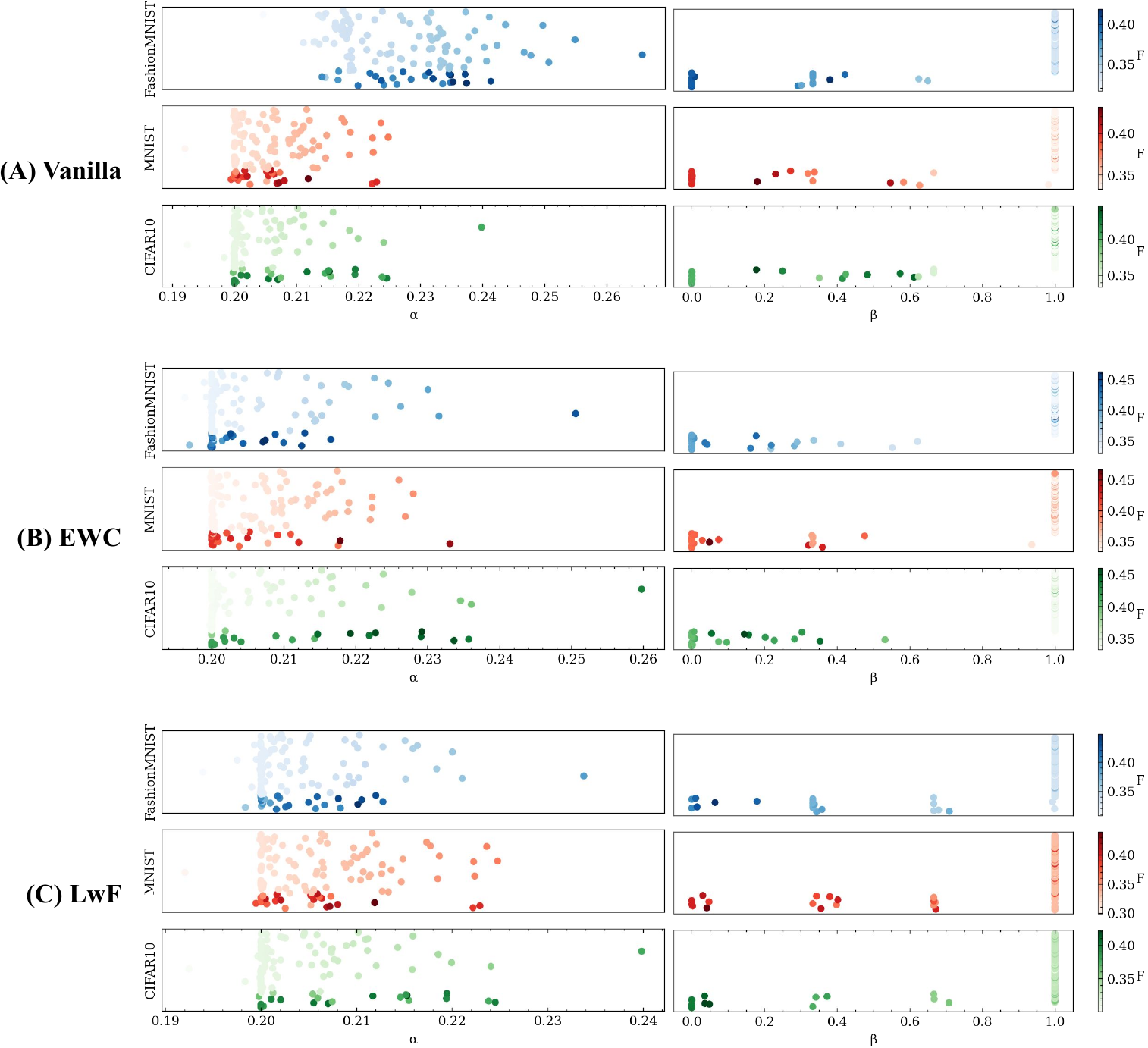}\vspace{-6mm}
\end{center}
  \caption[\textbf{Scatter plots showing how curriculum affects learning performance of the Vanilla, EWC, and LwF CL algorithms across MNIST, FashionMNIST, and CIFAR10 in paradigm-I}]{ \textbf{Curriculum affects learning performance of the (A) Vanilla, (B) EWC, and (C) LwF CL algorithms (Sec~\ref{sec:algos}) across three datasets: MNIST, FashionMNIST, and CIFAR10 (Sec~\ref{sec:datasets}) in paradigm-I (5 classes, Sec~\ref{sec:datasets}).} 
  Note that the y-axis does not carry any meaning. All the dots are randomly spread along the y-axis for easy visualization of the $\alpha$ and $\beta$ distributions. 
  This figure uses the same design conventions as \textbf{Fig~\ref{fig:fig2}}. 
  }
\label{fig:fig.f.5}
\end{figure*}

\begin{figure*}[htbp]
\begin{center}
\includegraphics[width=0.8\textwidth]{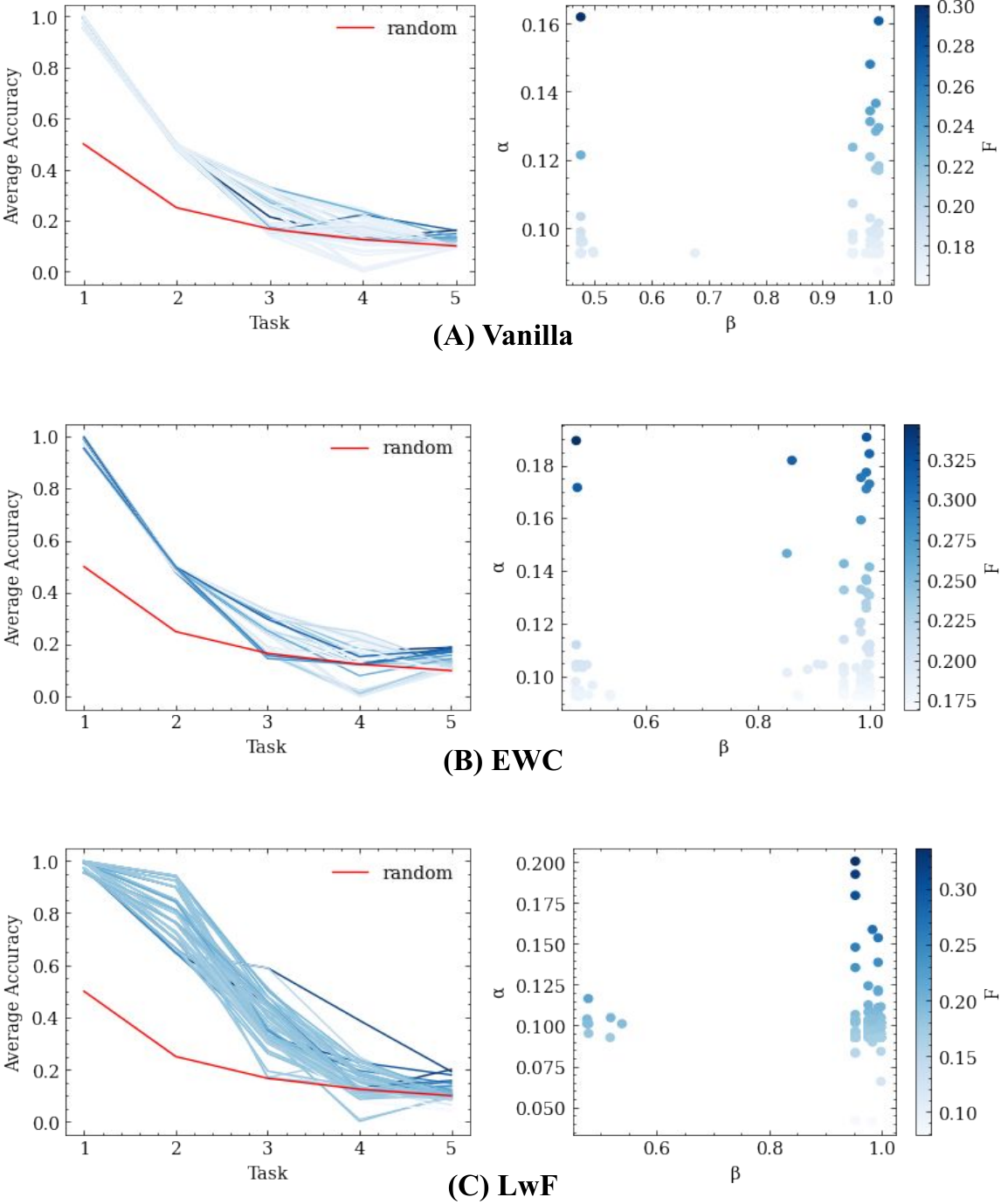}\vspace{-6mm}
\end{center}
  \caption[\textbf{Curriculum affects performance on MNIST of the Vanilla, EWC and LwF CL algorithms in paradigm-II}]{ \textbf{Curriculum affects performance on MNIST of the Vanilla, EWC and LwF CL algorithms (Sec~\ref{sec:algos}) in paradigm-II (10 classes arranged in 5 binary tasks, Sec~\ref{sec:datasets}).} This figure follows the same design conventions as \textbf{Fig~\ref{fig:fig.line.novelnet}}. 
  }
\label{fig:fig.line.mnist.10}
\end{figure*}

\begin{figure*}[htbp]
\begin{center}
\includegraphics[width=0.8\textwidth]{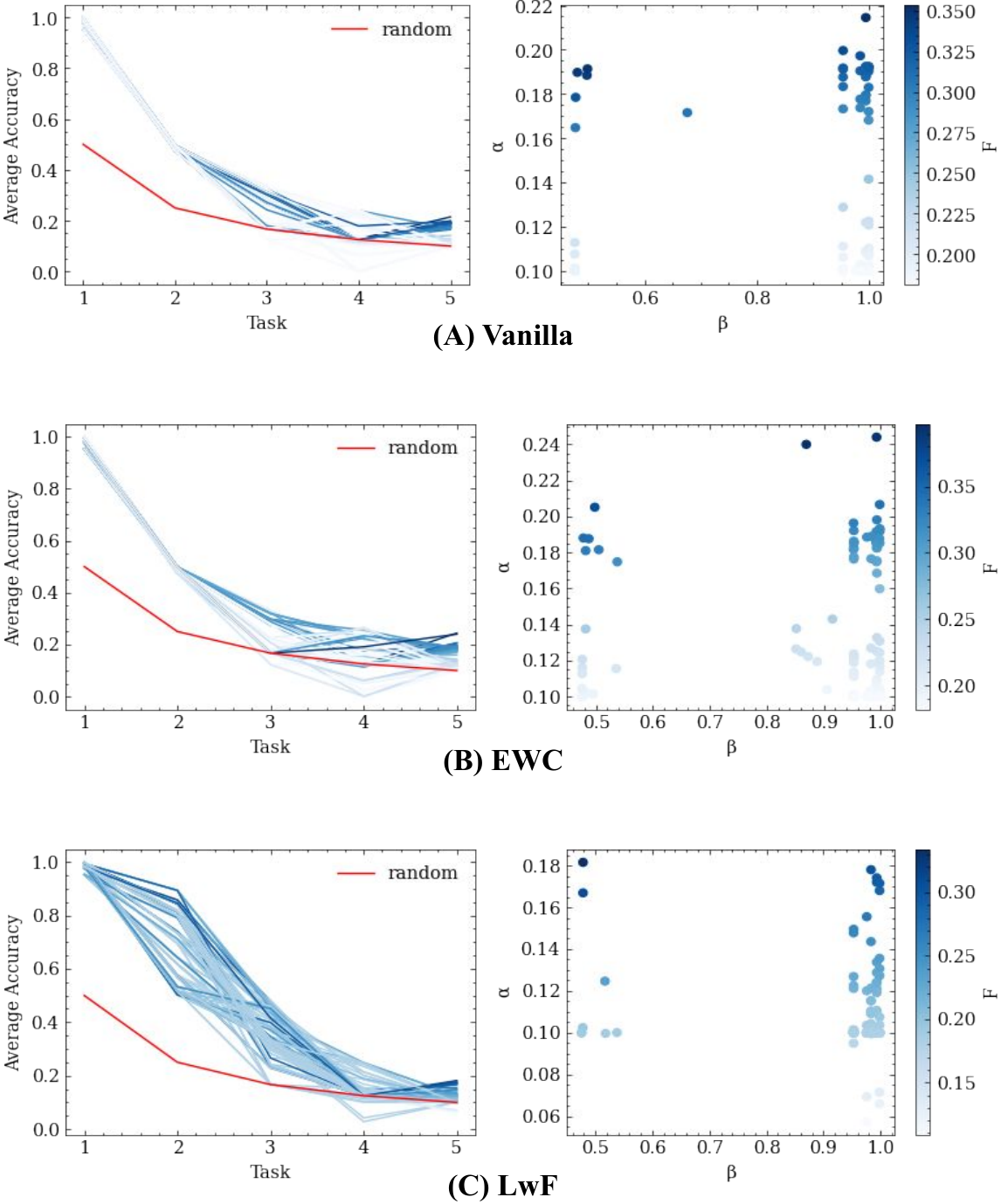}\vspace{-6mm}
\end{center}
  \caption[\textbf{Curriculum affects performance on FashionMNIST of the Vanilla, EWC and LwF CL algorithms in paradigm-II}]{ \textbf{Curriculum affects performance on FashionMNIST of the Vanilla, EWC and LwF CL algorithms (Sec~\ref{sec:algos}) in paradigm-II (10 classes arranged in 5 binary tasks, Sec~\ref{sec:datasets}).} This figure follows the same design conventions as \textbf{Fig~\ref{fig:fig.line.novelnet}}.
  }
\label{fig:fig.line.fmnist.10}
\end{figure*}

\begin{figure*}[htbp]
\begin{center}
\includegraphics[width=0.8\textwidth]{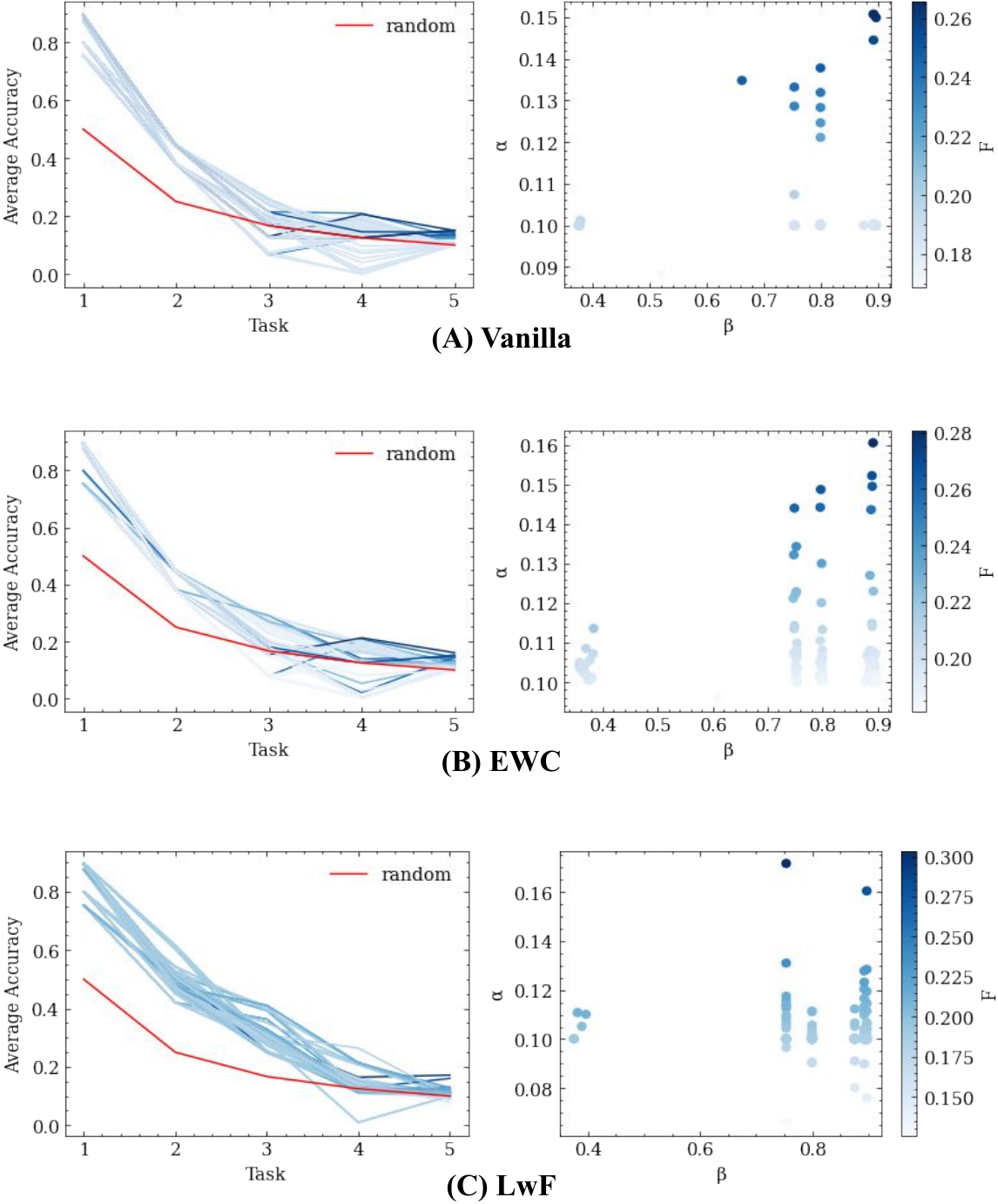}\vspace{-6mm}
\end{center}
  \caption[\textbf{Curriculum affects performance on CIFAR10 of the Vanilla, EWC and LwF CL algorithms in paradigm-II}]{ \textbf{Curriculum affects performance on CIFAR10 of the Vanilla, EWC and LwF CL algorithms (Sec~\ref{sec:algos}) in paradigm-II (10 classes arranged in 5 binary tasks, Sec~\ref{sec:datasets}).} This figure follows the same design conventions as \textbf{Fig~\ref{fig:fig.line.novelnet}}.
  }
\label{fig:fig.line.cifar.10}
\end{figure*}

\begin{figure*}[htbp]
\begin{center}
\includegraphics[width=\textwidth]{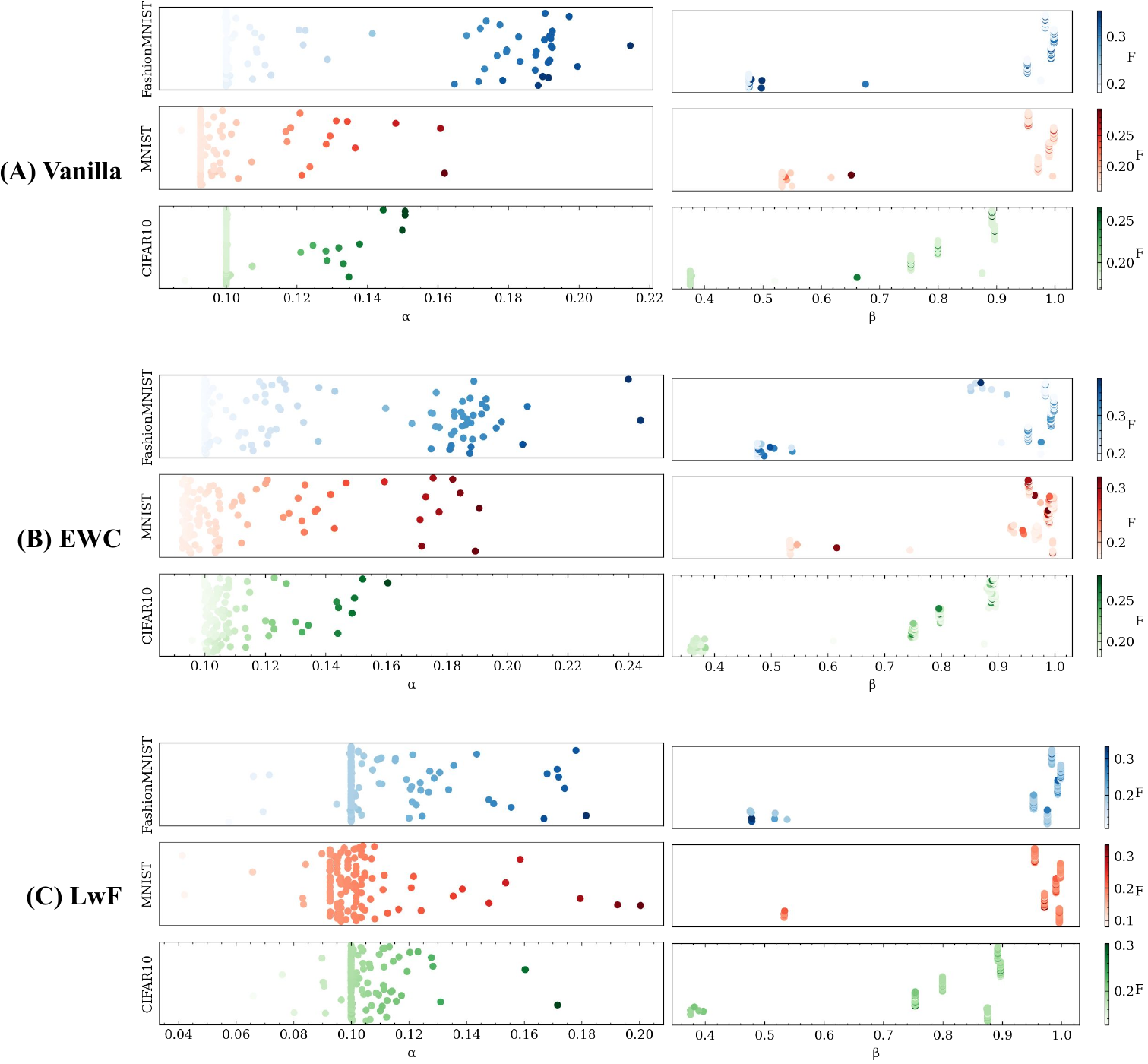}\vspace{-6mm}
\end{center}
    \caption[\textbf{Scatter plots showing how curriculum affects learning performance of the Vanilla, EWC, and LwF CL algorithms across MNIST, FashionMNIST, and CIFAR10 in paradigm-II}]{ \textbf{Curriculum affects learning performance of the (A) Vanilla, (B) EWC, and (C) LwF CL algorithms (Sec~\ref{sec:algos}) across three datasets: MNIST, FashionMNIST, and CIFAR10 (Sec~\ref{sec:datasets}) in paradigm-II (10 classes arranged in 5 binary tasks, Sec~\ref{sec:datasets}).} 
    Note that the y-axis does not carry any meaning. All the dots are randomly spread along the y-axis for easy visualization of the $\alpha$ and $\beta$ distributions. This figure uses the same design conventions as \textbf{Fig~\ref{fig:fig2}}. 
  }
\label{fig:fig.f.10}
\end{figure*}

\begin{figure*}[htbp]
\begin{center}
\includegraphics[scale=0.6]{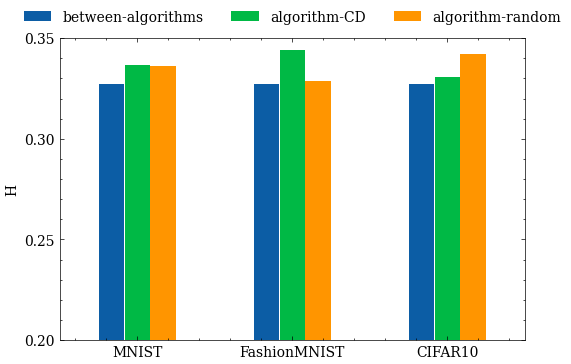}\vspace{-2mm}
  \caption[\textbf{Like in paradigm-I, in paradigm-II there is agreement among methods in ranking curricula by effectiveness}]{\textbf{Like in paradigm-I, in paradigm-II there is agreement among methods in ranking curricula by effectiveness.} Different CL algorithms agree with each other on which curricula are more effective than others, and also with our CD's heuristic estimates of relative curriculum optimality. Curriculum discrepancy $\mathcal{H}$ (\textbf{Sec~\ref{sec:eval}}) is reported between pairs of CL algorithms (between-algorithms, blue, averaging across all pairs), between CL algorithms and our CD (algorithm-CD, green, averaging across CL algorithms), and between CL algorithms and the random CD (algorithm-random, orange, averaging across CL algorithms) across MNIST, FashionMNIST, and CIFAR10 (\textbf{Sec~\ref{sec:curricula_agreement}}). See \textbf{Sec~\ref{sec:curricula_agreement}} for an analysis of these results. 
  }\vspace{-5mm} 
\label{fig:fig.curricula.agreement}
\end{center}
\end{figure*}

\begin{figure*}[htbp]
\begin{center}
\includegraphics[width=1\textwidth, scale=1]{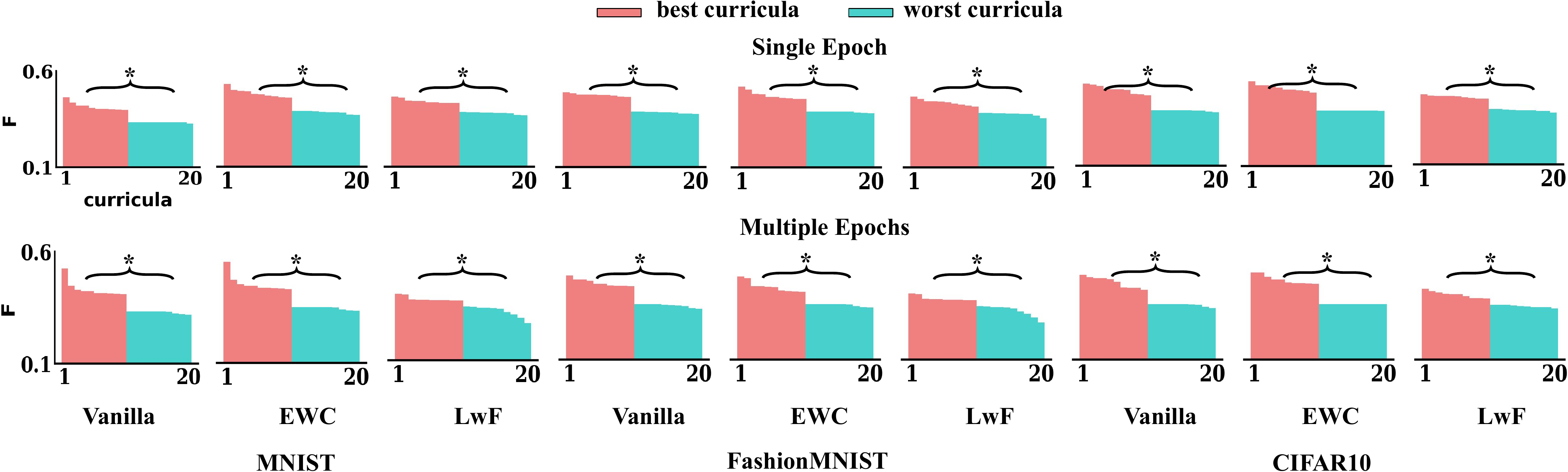}\vspace{-2mm}
  \caption[\textbf{Top 10 vs bottom 10 curricula across three datasets and three CL algorithms in paradigm-I}]{\textbf{Top 10 vs bottom 10 curricula across three datasets and three CL algorithms (Sec~\ref{sec:algos}) in paradigm-I (5 classes, Sec~\ref{sec:datasets}).} The top row of plots shows the online setting (single epoch per task), and the bottom row shows the offline setting with multiple epochs per task. Each plot shows the $\mathcal{F}$ scores for the best 10 curricula (red) and the worst 10 curricula (blue). The statistical significance (* = statistically significant) was determined using two-sample t-tests on the 10 best and 10 worst $\mathcal{F}$ scores. See \textbf{Sec~\ref{sec:curriculum_impact}}, and \textbf{\ref{sec:multiple_epochs}} for results on the impact of curricula in paradigm-I. 
  }\vspace{-5mm} 
\label{fig:fig.top10_bot10_paradigmI}
\end{center}
\end{figure*}

\begin{figure*}[htbp]
\begin{center}
\includegraphics[width=1\textwidth, scale=1]{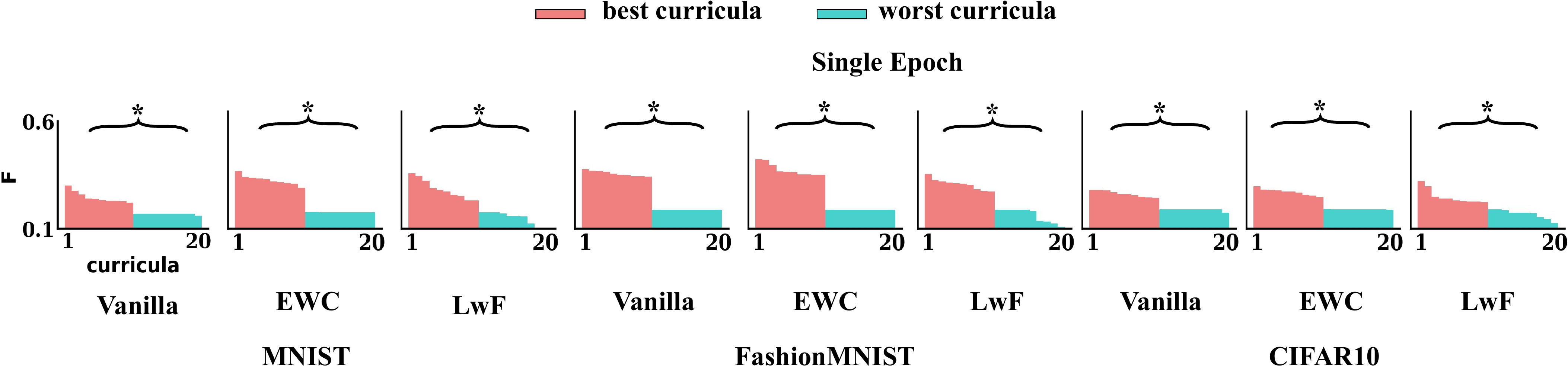}\vspace{-2mm}
  \caption[\textbf{Top 10 vs bottom 10 curricula across three datasets and three CL algorithms in paradigm-II}]{ \textbf{Top 10 vs bottom 10 curricula across three datasets and three CL algorithms (Sec~\ref{sec:algos}) in paradigm-II (10 classes arranged into 5 binary tasks, Sec~\ref{sec:datasets}).} See \textbf{Fig~\ref{fig:fig.top10_bot10_paradigmI}} for figure design conventions. See \textbf{Sec~\ref{sec:curriculum_impact}} for results on the impact of curricula in paradigm-II.
  }\vspace{-5mm} 
\label{fig:fig.top10_bot10_paradigmII}
\end{center}
\end{figure*}

\begin{figure*}[htbp]
\begin{center}
\includegraphics[width=0.6\textwidth, scale=1]{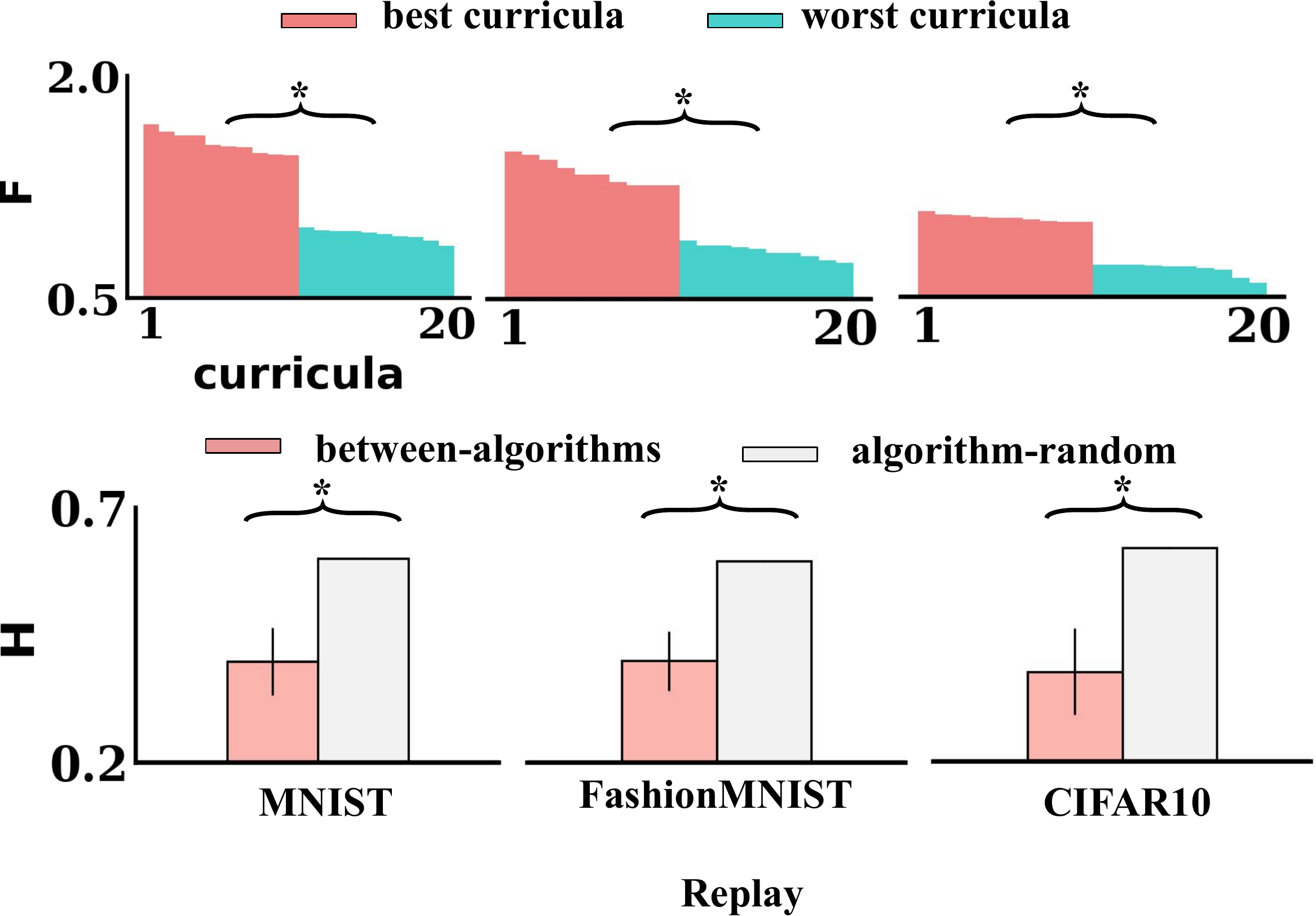}\vspace{-2mm}
  \caption[\textbf{Top 10 vs bottom 10 curricula, and curriculum discrepancy $\mathcal{H}$, for a naive replay CL algorithm across three datasets in paradigm-I}]{ \textbf{Top 10 vs bottom 10 curricula, and curriculum discrepancy $\mathcal{H}$ (Sec~\ref{sec:eval}), for a naive replay CL algorithm 
  across three datasets in paradigm-I (5 classes, Sec~\ref{sec:datasets})}. See \textbf{Sec.\ref{sec:algos}} for the introduction to the naive replay CL algorithm.
  Each plot in the top row shows the $\mathcal{F}$ scores for the best 10 curricula (red) and the worst 10 curricula (blue). The second row shows curriculum agreement plots for each dataset (see \textbf{Sec~\ref{sec:replay}} for details). Statistical significance (* = statistically significant) was determined using two-sample t-tests between the 10 highest and 10 lowest $\mathcal{F}$ or $\mathcal{H}$ scores (\textbf{Sec~\ref{sec:stat_anal}}). 
  The errorbars are the standard deviations over all the test trials. The errorbars are small; and hence, they become almost invisible. See \textbf{Sec.\ref{sec:stat_anal}}
for statistical interpretations and analysis.  
  }\vspace{-5mm} 
\label{fig:fig.replay}
\end{center}
\end{figure*}

\begin{figure*}[htbp]
\begin{center}
\includegraphics[width=0.8\textwidth, scale=1]{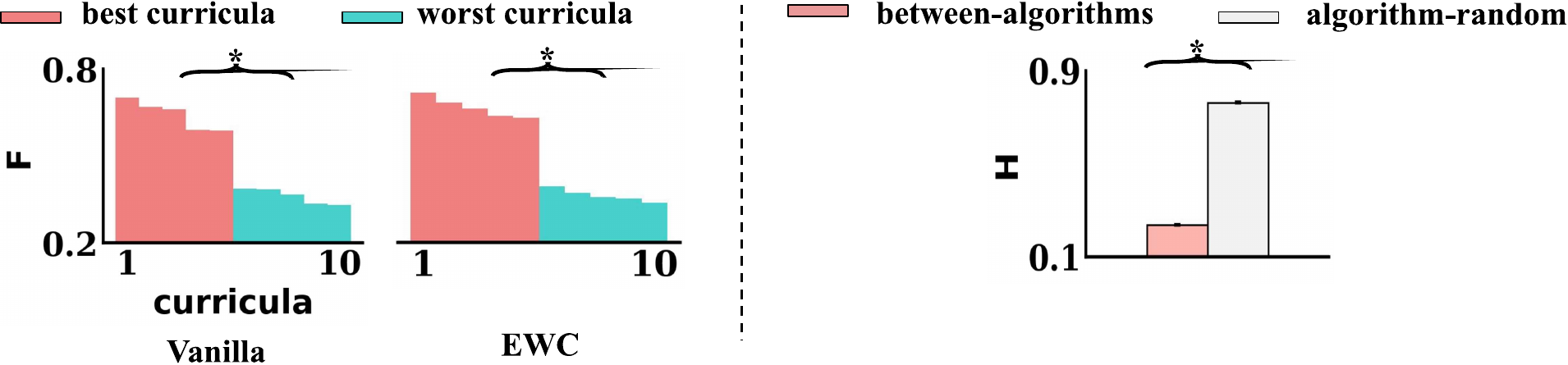}\vspace{-2mm}
  \caption[\textbf{Strong curriculum effects are observed in the continual visual question answering setting}]{ \textbf{Strong curriculum effects are observed in the continual visual question answering (VQA) setting.} 
  The left panel shows $\mathcal{F}$ scores for the best 5 and worst 5 curricula using the Vanilla and EWC CL algorithms.  and the curricula agreement $\mathcal{H}$ plot for continual VQA (\textbf{Sec~\ref{sec:vqa}}) on the CLOVER dataset~\cite{lei2022symbolic}. See \textbf{Sec~\ref{sec:vqa}} for further analysis and details of VQA experiments. 
  }\vspace{-5mm} 
\label{fig:fig.vqa}
\end{center}
\end{figure*}

\begin{figure*}[htbp]
\begin{center}
\includegraphics[width=0.6\textwidth, scale=1]{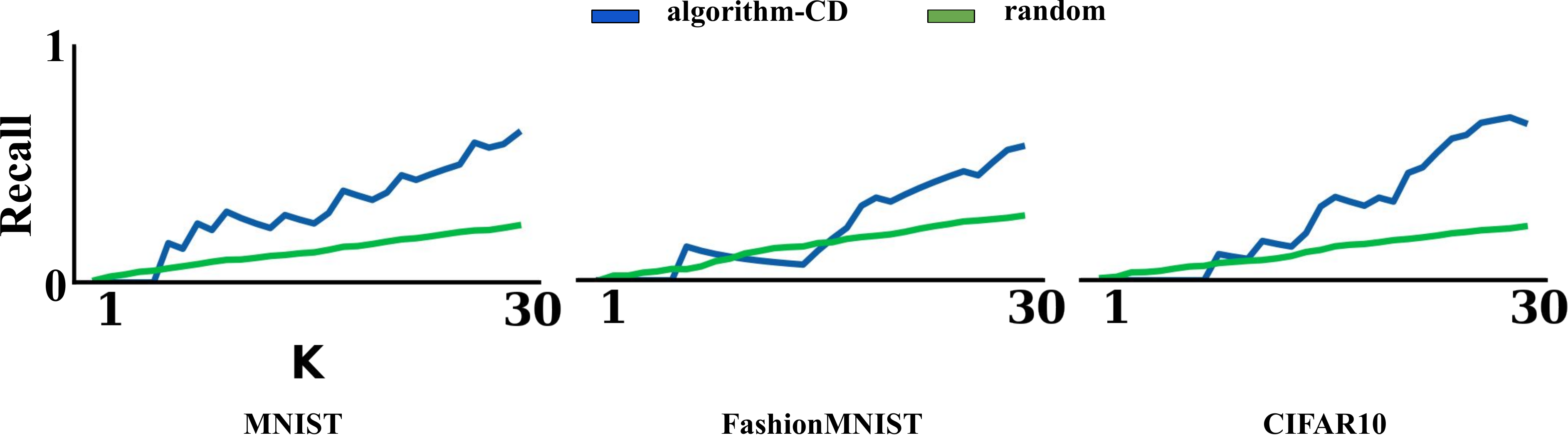}\vspace{-2mm}
  \caption[\textbf{Our curriculum designer (CD) predicts optimal curricula more accurately than a random CD in paradigm-II}]{ 
  \textbf{Our curriculum designer (CD) predicts optimal curricula more accurately than a random CD in paradigm-II (10 classes arranged in 5 binary tasks, \textbf{Sec~\ref{sec:datasets}).}} See \textbf{Fig~\ref{fig:fig3}} for the equivalent plots for paradigm-I. The results are analysed in \textbf{Sec~\ref{sec:optimal_curricula}.}
  }\vspace{-5mm} 
\label{fig:fig.recall@allk_10}
\end{center}
\end{figure*}

\begin{figure*}[htbp]
\begin{center}
\includegraphics[width=\textwidth]{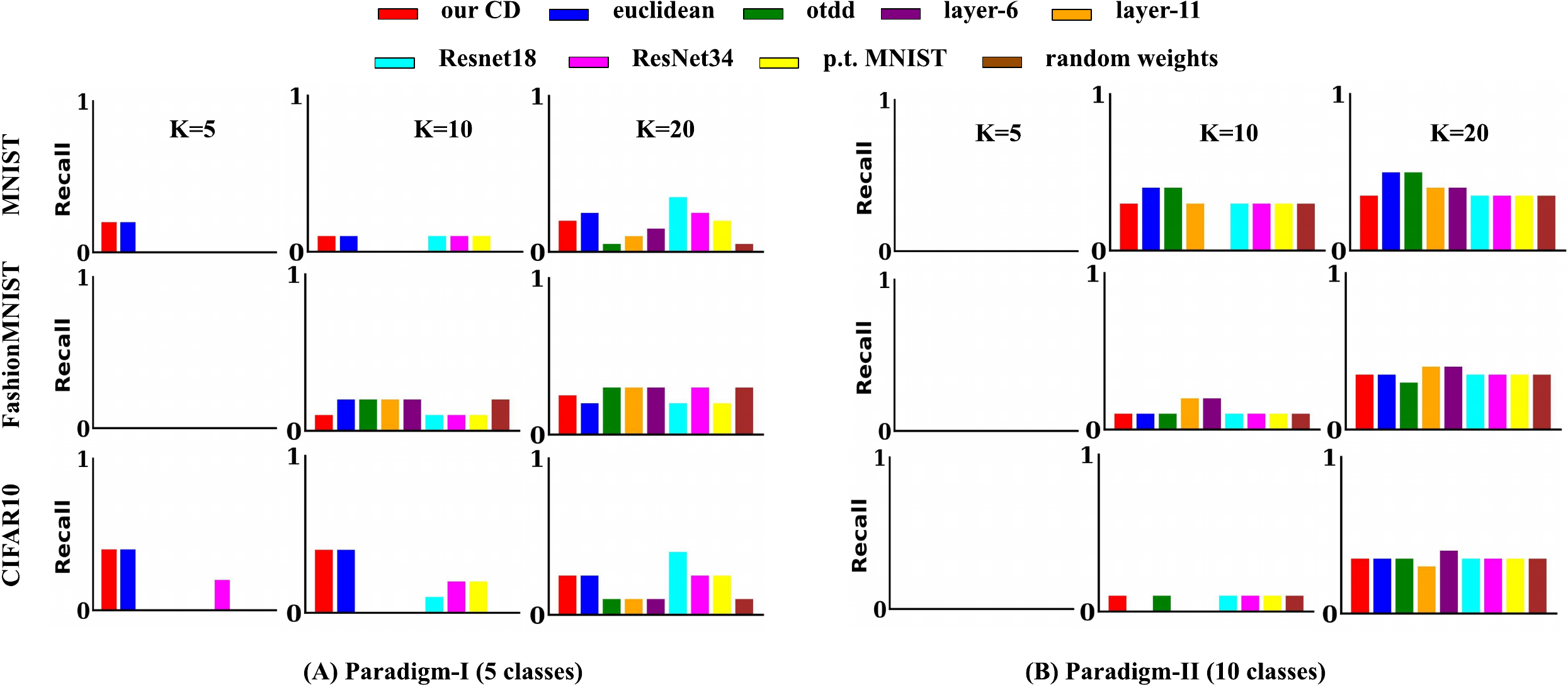}\vspace{-2mm}
  \caption[\textbf{Ablation study results on our CD in paradigms I and II}]{ \textbf{Ablation study results on our CD in (A) paradigm-I (5 classes, Sec~\ref{sec:datasets}) and (B) paradigm-II (10 classes arranged in 5 binary tasks, Sec~\ref{sec:datasets}).} See \textbf{Fig~\ref{fig:fig4}} for the same design convention. The results in paradigm-I are analyzed in \textbf{Sec~\ref{sec:results.3}}, and the results in paradigm-II are analyzed in \textbf{Sec~\ref{sec:cd.design}}.
  }\vspace{-5mm} 
\label{fig:fig.recall@k}
\end{center}
\end{figure*}

\begin{figure*}[htbp]
\begin{center}
\includegraphics[width=0.8\textwidth, scale=1]{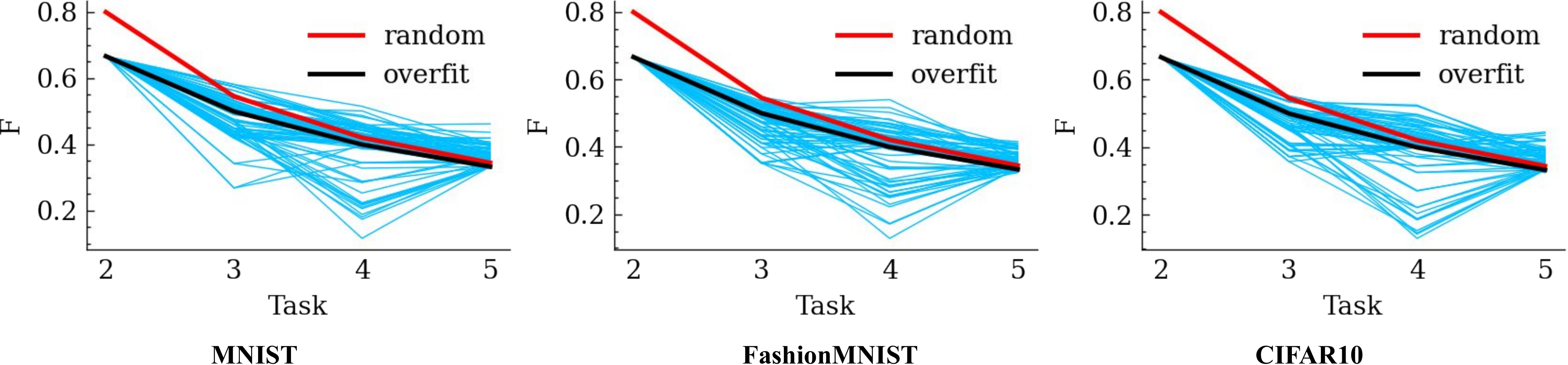}\vspace{-2mm}
  \caption[\textbf{Task-wise $\mathcal{F}$ of the Vanilla CL algorithm across three datasets in paradigm-I}]{
  \textbf{Task-wise $\mathcal{F}$ of the Vanilla CL algorithm across three datasets in paradigm-I (5 classes, Sec~\ref{sec:datasets}).} Task-wise $\mathcal{F}$ is shown as a blue line plot for each of the $5!=120$ possible curricula on MNIST, FashionMNIST and CIFAR10. The performance on each task of the random model (red) and a completely over-fitting CL algorithm (black) are also shown (\textbf{Sec~\ref{sec:f_overtime}}). See \textbf{Sec~\ref{sec:f_overtime}} for the baseline introductions.
  }\vspace{-5mm} 
\label{fig:fig.f_overtime}
\end{center}
\end{figure*}

\clearpage

\begin{algorithm}[t]
\footnotesize
\SetAlgoLined
    \PyComment{N: number of classes}\\
    \PyComment{M (N $\times$ N): M[i][j] is the distance between the feature prototypes of class i and class j; 
    } \\
    \PyComment{Var(): function to compute variance} \\
    \PyComment{C: a given curriculum in sequence of C[1], C[2],...C[i],...,C[N], where i is the class index}\\
    \PyComment{initialize ranking score s} \\
    \PyCode{s = 0} \\
    \PyComment{at i = 1} \\
    \PyCode{s = 1 - Var({M[1][j]}$^{N}_{j=2}$)} \\
    \PyCode{for t in (2, N):} \\
    \Indp
        \PyCode{if t $\leq$ ($\lfloor \frac{N}{2} \rfloor$)}\\
            \Indp   
                \PyCode{s += M[t][t-1]} \\
            \Indm
    \Indm
    \Indp
        \PyCode{if t > ($\lfloor \frac{N}{2} \rfloor$)}\\
            \Indp   
                \PyCode{s += 1 - M[t][N - t + 1]} \\
            \Indm
    \Indm
\caption{Python-style pseudocode for CD}
\label{algo:your-algo}
\end{algorithm}

\clearpage
{\small
\bibliographystyle{ieee_fullname}
\bibliography{egbib}
}


\end{document}